\PassOptionsToPackage{svgnames,table,prologue,dvipsnames}{xcolor}
\documentclass[10pt,twocolumn,letterpaper]{article}
\pdfoutput=1

\usepackage[table]{xcolor}
\usepackage{color}
\newcommand{\mygray}{\rowcolor[HTML]{f5f5f5}}
\newcommand{\bestcolor}[1]{\textbf{\color[HTML]{C00000} #1}}
\newcommand{\secondcolor}[1]{\textbf{\color[HTML]{0070C0} #1}}

\definecolor{darkgreen}{rgb}{0.0, 0.4, 0.3}
\newcommand{\CommentGreen}[1]{\textcolor{darkgreen}{\# #1}}

\usepackage{iccv}              

\definecolor{iccvblue}{rgb}{0.21,0.49,0.74}

\usepackage{lineno}
\usepackage{pifont}
\usepackage{graphicx}
\usepackage{listings}
\usepackage[pagebackref,breaklinks,colorlinks,allcolors=iccvblue]{hyperref}
\usepackage{bbding}
\usepackage{algorithm}
\usepackage{algorithmic}
\usepackage{amsmath}
\usepackage{makecell}
\usepackage{amssymb}
\usepackage{bbm}
\usepackage{booktabs}
\usepackage{mathrsfs}
\usepackage[percent]{overpic}  
\usepackage{multirow}

\usepackage{soul}
\usepackage[normalem]{ulem}

\def\paperID{2410} 
\def\confName{ICCV}
\def\confYear{2025}
\def\ourmodel{LawDIS}

\title{LawDIS: Language-Window-based \\Controllable Dichotomous Image Segmentation}

\author{Xinyu Yan$^{1,2,6}$ \quad Meijun Sun$^{1,2}$ \quad Ge-Peng Ji$^{3}$ \\ Fahad Shahbaz Khan$^{6}$ \quad Salman Khan$^{6}$ \quad Deng-Ping Fan$^{4,5}$\thanks{Corresponding author: Deng-Ping Fan (dengpfan@gmail.com)}\\
\small $^{1}$ Tianjin University \quad $^2$ Tianjin Key Laboratory of Machine Learning \quad $^3$ Australian National University \\
\small $^4$ Nankai Institute of Advanced Research (SHENZHEN FUTIAN) \quad $^5$ Nankai University \quad $^6$ MBZUAI 
}

\begin{document}

\maketitle

\begin{abstract}
\vspace{-12pt}

We present \ourmodel, a language-window-based controllable dichotomous image segmentation (DIS) framework that produces high-quality object masks. Our framework recasts DIS as an image-conditioned mask generation task within a latent diffusion model, enabling seamless integration of user controls. \ourmodel~is enhanced with macro-to-micro control modes. Specifically, in macro mode, we introduce a language-controlled segmentation strategy (LS) to generate an initial mask based on user-provided language prompts. In micro mode, a window-controlled refinement strategy (WR) allows flexible refinement of user-defined regions (i.e., size-adjustable windows) within the initial mask. Coordinated by a mode switcher, these modes can operate independently or jointly, making the framework well-suited for high-accuracy, personalised applications. Extensive experiments on the DIS5K benchmark reveal that our \ourmodel~significantly outperforms 11 cutting-edge methods across all metrics. Notably, compared to the second-best model MVANet, we achieve $F_\beta^\omega$ gains of 4.6\% with both the LS and WR strategies and 3.6\% gains with only the LS strategy on DIS-TE. Codes will be made available at \url{https://github.com/XinyuYanTJU/LawDIS}.

\end{abstract}

\begin{figure}[t!]
\centering
\begin{overpic}[width=0.48\textwidth]{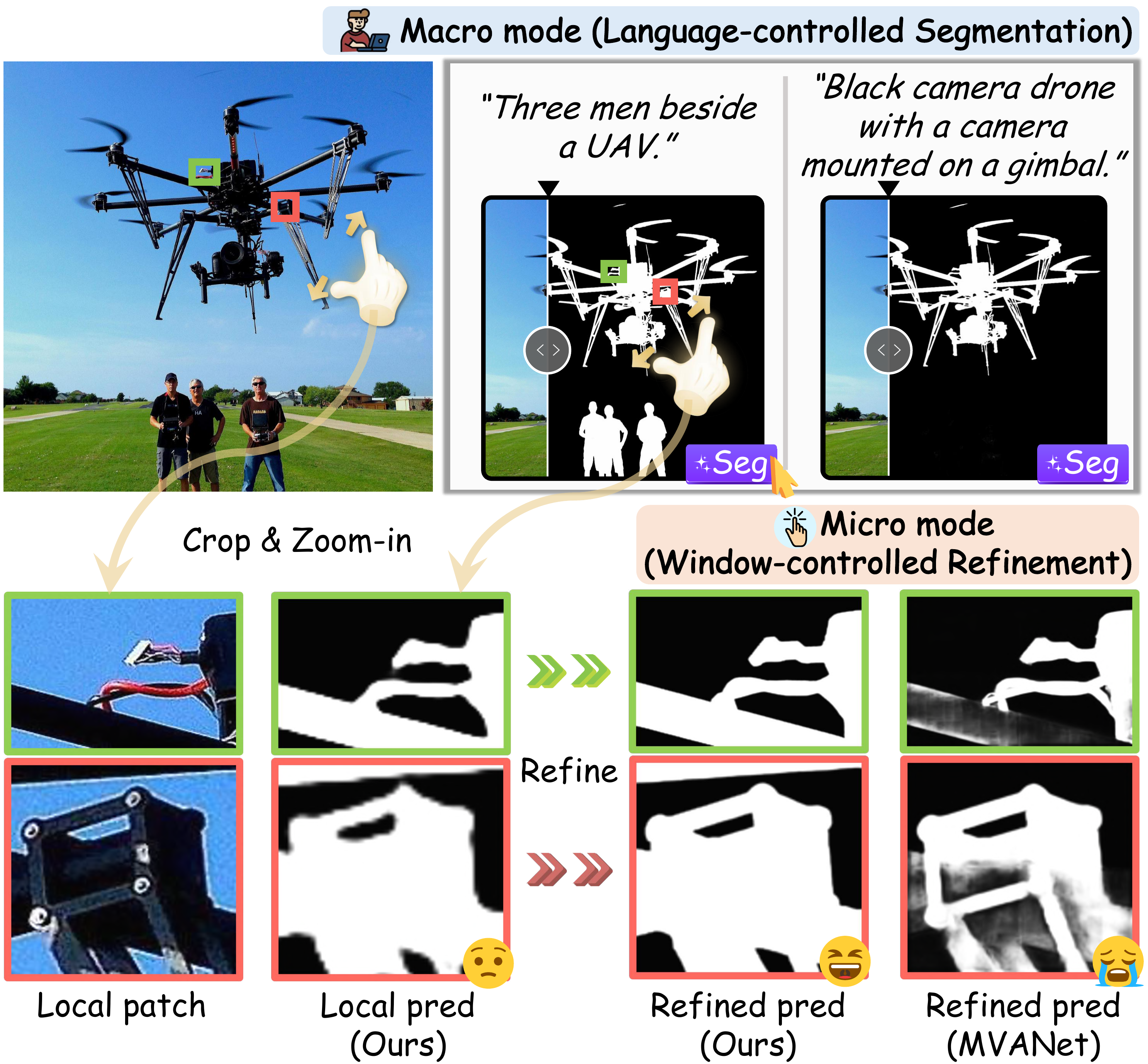}
\end{overpic}
\vspace*{-15pt}
\caption{
Illustration of the proposed macro and micro controls for the DIS task. The macro mode enables users to segment objects with customised language prompts, while micro mode supports post-refinement on user-defined windows at any scale. After refinement, our results become more precise, whereas the runner-up model, MVANet \cite{yu2024multi}, lacks adaptability to the cropped local patches, resulting to inferior predictions.
}
\label{fig:teaser}
\end{figure}
\vspace*{-10pt}


\section{Introduction}
\label{sec:intro}
\vspace{-5pt}
With the popularity of high-quality camera equipment, segmentation task in computer vision has evolved from rough localisation \cite{treisman1980feature,MIR-2023-09-178} to high-precision delineation \cite{zeng2019towards}. As a result, the dichotomous image segmentation (DIS) task \cite{qin2022highly} -- focused on segmenting highly accurate foreground object(s) -- has garnered significant attention due to its broad range of applications, \eg, 3D reconstruction \cite{liu2021fully,tu2024overview}, image editing \cite{goferman2011context,liu2024referring}, augmented reality \cite{qin2021boundary,ying2024omniseg3d}, and medical image segmentation \cite{ji2022video,ji2024frontiers}.

Compared to general segmentation tasks \cite{lin2014microsoft, liu2024primitivenet,song2024vitgaze}, the DIS task presents greater challenges in two key aspects. First, the widely recognised DIS benchmark \cite{qin2022highly} covers more than 200 object categories and diverse visual traits, such as salient \cite{fan2022salient,zhuge2022salient} and camouflaged \cite{fan2021concealed,fan2023advances,ji2023gradient} objects. This diversity requires models to exhibit strong holistic understanding to accurately identify foreground objects across varied scenarios. Additionally, this task emphasizes highly precise object delineation in high-resolution images, capturing even the internal details of objects. Therefore, models should exhibit fine-grained feature extraction capabilities to effectively segment complex structures and shapes.

Current DIS methods \cite{qin2022highly,zhou2023dichotomous,pei2023unite,kim2022revisiting,zheng2024birefnet,yu2024multi} often adopt a discriminative learning paradigm (\ie, per-pixel classification) that depends on their intrinsic learning ability to derive object semantics. This paradigm would struggle in real-world applications, particularly when addressing personalised needs. First, when an image contains multiple foreground entities, it is semantically ambiguous which object(s) should be targeted. Second, to better capture geometric details of high-resolution objects, most methods \cite{pei2023unite,kim2022revisiting,xie2022pyramid} incorporate an extra high-resolution data flow, often by downsampling a full-size image to a $1024^2$px input. This straightforward way compensates for fine-grained details of a segmented object, while infinitely scaling up the input size is computationally impractical. Recent methods \cite{yu2024multi,zheng2024birefnet} address this by splitting the full-size image into a series of patches, equivalently enlarging geometric details with fewer pixel losses. However, as these methods are trained on patches with predefined size, they lack adaptability to variable patch sizes. As shown in Fig.~\ref{fig:teaser}, the advanced model, MVANet \cite{yu2024multi}, fails when given a local patch of input size that differs from those used during training.

To alleviate the above issues, we present a language-windows-based controllable DIS framework, \ourmodel, designed to meet user-personalized requirements. Our framework recasts the DIS as an image-conditioned mask generation task within a latent diffusion model \cite{rombach2022high}, enabling the seamless integration of various user controls. Furthermore, we enhance this generative framework with macro-to-micro control modes. In macro mode, we propose a language-controlled segmentation strategy (LS) that generates an initial mask based on a user-provided language prompt, as shown in the upper part of Fig.~\ref{fig:teaser}. In micro mode, the window-controlled refinement strategy (WR) refines user-specified regions at variable scales based on the initial mask, as illustrated on the lower part of Fig.~\ref{fig:teaser}. These two modes can operate independently or jointly through a mode switcher, facilitating adaptive optimisation during training and mutual refinement during inference. Extensive experiments on the DIS5K benchmark \cite{qin2022highly} reveals that \ourmodel~significantly surpasses 11 state-of-the-art (SOTA) methods across all metrics. Specifically, compared to the second-best model, MVANet, we achieve an improvement of 4.6\% in $F_\beta^\omega$ when employing both LS and WR strategies, and an improvement of 3.6\% with the only LS strategy.

Our main contributions are summarized as follows: 
\begin{itemize}
\item We present \ourmodel, a language-window-based controllable framework that reformulates DIS as a image-conditioned mask generation problem, enabling seamless user-controlled integration for producing high-quality object masks.
\item This framework is enhanced with macro-to-micro control modes. In macro mode, we introduce a language-controlled segmentation strategy (LS) to generate an initial mask based on the user-provided language prompt. In micro mode, we design a window-controlled refinement strategy (WR) to refine user-specified regions (\ie, size-adjustable windows) within this initial mask.
\item These two modes can operate independently or jointly via a mode switcher, resulting in SOTA performance across all metrics on the DIS5K benchmark.
\end{itemize}

\section{Related Works}
\label{sec:related}

\noindent\textbf{Dichotomous image segmentation.} Qin~\etal \cite{qin2022highly} introduce the DIS task for accurately segmenting objects with varying structural complexities in high-resolution images, regardless of their characteristics, which has garnered significant interest from the research community. Following the advent of the fully convolutional network \cite{long2015fully}, many models formulate this task within a discriminative framework, treating it as a per-pixel classification problem. Early solutions \cite{qin2019basnet,qin2020u2,wang2020deep,qin2022highly,zhou2023dichotomous,pei2023unite} rely on convolutional designs, employing strategies such as intermediate supervision \cite{qin2022highly}, frequency priors \cite{zhou2023dichotomous}, and unite-divide-unite \cite{pei2023unite} to improve performance. Recently, visual transformers \cite{dosovitskiy2020image} have gained favour due to their powerful capability to model long-range dependencies, resulting in impressive results.  However, due to the absence of convolutional local inductive bias, their ability to capture local structures remains relatively weak. To overcome this limitation, InSPyReNet \cite{kim2022revisiting}, BiRefNet \cite{zheng2024birefnet}, and MVANet \cite{yu2024multi} introduce additional images or patches of varying resolutions as input during the training or inference stages, enriching the capture of detailed information to some extent.

The aforementioned DIS methods, which rely on discriminative learning paradigms, primarily focus on balancing and integrating global and local information but tend to overlook two key challenges: flexible semantic control across different scenarios and local window refinement for unavoidable blurry segmentation. Different from these methods, the proposed \ourmodel~reformulates the DIS task within a generative-based model, leveraging the encyclopedic visual-language understanding and the advantages of denoising mechanisms to achieve language-controlled segmentation and window-controlled refinement.

\noindent\textbf{Diffusion models for high-resolution segmentation.}
Currently, a trend in the computer vision community adapting latent diffusion model \cite{rombach2022high} to high-resolution segmentation tasks \cite{DiffDIS}. Considering the challenges associated with high-resolution data acquisition, a reasonable approach \cite{qianmaskfactory} is to leverage the powerful generative capabilities of stable diffusion to synthesize data, thereby enhancing the performance of existing high-resolution segmentation methods. In addition, Wang~\etal \cite{wang2023segrefiner} propose a refinement model to enhance the quality of object masks generated by different segmentation models using the diffusion process. Later, a two-stage latent diffusion approach \cite{wang2024MG} has employed  for portrait matting. GenPercept \cite{xu2024diffusion} transforms the generative model into a deterministic one-step fine-tuning paradigm through a customized decoder and conducts extensive experiments on a wide range of fundamental visual dense perception tasks, including high-resolution segmentation tasks such as DIS and portrait matting. Distinct from prior methods, we extend a single stable diffusion to both macro and micro modes using a switcher. This not only enables both high-resolution image segmentation and result refinement within the same model, but also allows user control over both language and window aspects.

\section{Methodology}
\label{sec:methd}
We propose a user-controlled framework via reformulating DIS task within an image-conditioned mask generation paradigm. Our framework supports two levels of control. Specifically, the macro-mode allows users to specify and segment the objects from the high-resolution image, guided by custom language prompts. Meanwhile, the micro-mode functions as a general post-refinement tool to improve the accuracy of segmentation masks, particularly in areas with intricate structures. For better integrability, we consolidate macro- and micro-modes within one diffusion model using a mode switcher, yielding better geometric representations at varied scales. Next, we introduce the preliminaries to build the generative paradigm (Sec.~\ref{sec:Generative Formulation}) and two key processes (Sec.~\ref{sec:Global Seg} \& Sec.~\ref{sec:Local Refine}) to repurpose it for the DIS task.

\subsection{Generative Formulation for DIS}
\label{sec:Generative Formulation}
We redefine the DIS task involving conditional denoising diffusion, focusing on modelling the conditional probability distribution $D(s\mid x)$ for segmentation masks $s \in \mathbb{R}^{W \times H}$. The condition $x \in \mathbb{R}^{W \times H \times 3}$ is specified by the corresponding RGB image awaiting segmentation. Utilizing the framework established by a pre-trained Stable Diffusion v2 model \cite{rombach2022high}, our \ourmodel~efficiently executes the diffusion process within a lower-dimensional latent space. Initially, a variational autoencoder with an encoder $\phi(\cdot)$ and a decoder $\varphi(\cdot)$ is used to transform data between segmentation mask and latent space, such that $\mathbf{z}^{(s)}=\phi(s)$ and $s\approx\varphi(\mathbf{z}^{(s)})$. Similarly, $x$ is transposed to latent representation as $\mathbf{z}^{(x)}=\phi(x)$, serving as a condition for generation.

Second, stable diffusion sets up a \textit{forward} nosing process and a \textit{reversal} denoising process within the latent space using a U-Net. In \textit{forward} process, starting from $\mathbf{z}_{0}^{(s)}:=\mathbf{z}^{(s)}$, Gaussian noise $\boldsymbol{\epsilon} \sim \mathcal{N}(0, I)$ is gradually added at each level $t \in\{1, \ldots, T\}$ to construct a discrete Markov chain $\{\mathbf{z}_{0}^{(s)}, \mathbf{z}_{1}^{(s)}, \ldots, \mathbf{z}_{T}^{(s)}\}$.

\begin{figure}[t!]
\centering
\begin{overpic}[width=0.48\textwidth]{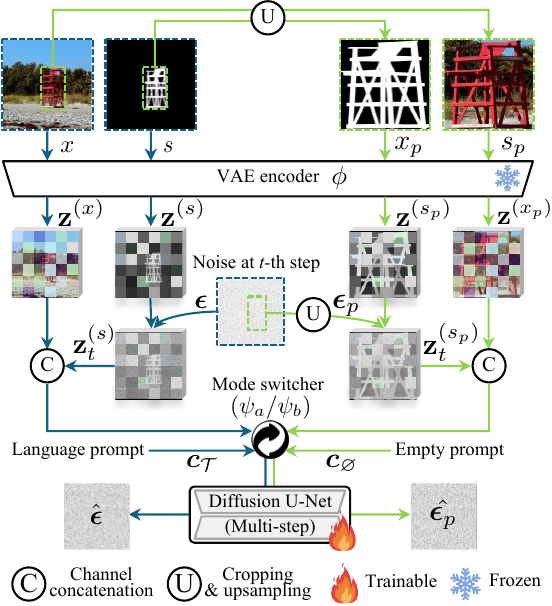}
\end{overpic}
\vspace*{-10pt}
\caption{An overview of the U-Net training protocol in \ourmodel. Starting from the pretrained stable diffusion, we introduce a mode switcher to extend it to both macro and micro modes for joint training, where $\psi_a$ activates macro mode and $\psi_b$ activates micro mode. The entire image $x$ and its segmentation map $s$ serve as inputs for the macro mode, while its patch image $x_p$ and patch map $s_p$ serve as inputs for the micro mode. Both sets of input are fed into the VAE encoder, converting them into latent space. Paired with language prompt $\boldsymbol{c}_{\mathcal{T}}$ and empty prompt $\boldsymbol{c}_{\varnothing}$ respectively, these latent representations are then fed into the U-Net, resulting in $\hat{\boldsymbol{\epsilon}}$ and $\hat{\boldsymbol{\epsilon}_{p}}$, which are sent for optimization of the standard diffusion objective relative to segmentation latent codes.}
\label{fig:train_test}
\end{figure}

In \textit{reversal} process, the U-Net model $f_{\theta}(\cdot)$ with learned parameters $\theta$, gradually predicts the noise $\boldsymbol{\epsilon}$ added to the noisy sample $\mathbf{z}_{t}$ at each step $t$. $\theta$ is updated as follows: a data pair $(s, x)$ is sampled from the training set and converted to $(\mathbf{z}^{(s)},\mathbf{z}^{(x)})$, noise $\boldsymbol{\epsilon}$ is added to $\mathbf{z}^{(s)}$ at a random time step $t$, and the estimated noise $\hat{\boldsymbol{\epsilon}}$ is computed as $f_{\theta}(\mathbf{z}_{t}^{(s)}, \mathbf{z}^{(x)}, t)$. The objective is to minimize:
\begin {equation} 
 \begin{aligned} 
\mathbb{E}_{\epsilon \sim \mathcal{N}(0,1), t, \mathbf{z}_{0}^{(s)}, \mathbf{z}^{(x)}}[\|\boldsymbol{\epsilon}-f_{\theta}(\mathbf{z}_{t}^{(s)}, \mathbf{z}^{(x)}, t)\|_{2}^{2}].
\end{aligned}
\label{equ:naiveloss}
\end {equation}

During inference, noise $\hat{\boldsymbol{\epsilon}}$ can be progressively predicted starting from a normally distributed variable $\mathbf{z}_{T}^{(s)} \in \mathcal{N}(\mathbf{0}, \mathbf{1})$, and then gradually denoised under the guidance of denoising schedulers (such as DDPM \cite{ho2020denoising}, DDIM \cite{song2020denoising}, \textit{etc.}) to obtain $\mathbf{z}_{0}^{(s)}$. Subsequently, the estimated clean latent representation $\mathbf{z}_{0}^{(s)}$ is reconstructed through the decoder $\varphi$ to get the mask prediction $\hat{s }= \varphi(\mathbf{z}_{0}^{(s)}) \sim D(s \mid x  )$.

\subsection{Joint Training with Two Modes}
\label{sec:Global Seg}
\noindent\textbf{Mode switcher.}
To enable \ourmodel~to perform different functions in two modes, we introduce a mode switcher $\psi$ to stable diffusion, which is represented as a one-dimensional vector encoded by the positional encoding and then added with the time embeddings of diffusion model. $\psi$ is set to either $\psi_a$ or $\psi_b$, which activates the macro mode or the micro mode, respectively. The two modes are designed to mutually enhance each other during training and enable seamless switching during inference.

\noindent\textbf{Macro mode.}
When $\psi_a$ is activated, \ourmodel~switches to the macro mode, where a language-controlled segmentation strategy is employed to segment objects with the guidance of the user-provided prompt. During training, the model receives the full image $x$, segmentation mask $s$ and its corresponding user prompts $\mathcal{T}$. These prompts are generated by VLM \cite{achiam2023gpt, chen2023minigpt}. See \hyperlink{supplementary_anchor}{\textcolor{magenta}{supp.}} for more details. As described in Sec.~\ref{sec:Generative Formulation}, $x$ and $s$ are encoded into latent space, which are then passed into the diffusion process to learn to predict the added Gaussian noise $\boldsymbol{\epsilon}$. As per \cite{rombach2022high}, we use CLIP \cite{radford2021learning} to encode prompts $\mathcal{T}$ into a control embedding $\boldsymbol{c}_{\mathcal{T}}$, which is integrated into the diffusion U-Net via cross-attention mechanism. The loss function for the macro mode is:
\begin {equation} 
 \begin{aligned} 
 \mathcal{L}_{macro}=\|\boldsymbol{\epsilon}-f_{\theta}(\mathbf{z}_{t}^{(s)}, \mathbf{z}^{(x)},\boldsymbol{c}_{\mathcal{T}}, t, \psi_a)\|_{2}^{2}.
\end{aligned}   
\end {equation}

\noindent\textbf{Micro mode.}
When $\psi_b$ is selected, \ourmodel~activates its micro mode, employing a window-controlled refinement strategy to precisely delineate details within user-specified windows. During training, we select the minimum enclosing rectangle of the foreground object in the segmentation mask $s$ as the local window, rather than using random windows. This ensures that the foreground object is fully within the window, thus preventing missing it or introducing ambiguous semantics. Based on this window selection criteria, we crop the corresponding regions from the image $x$ and the segmentation mask $s$, resulting in local patch $x_p$ and local mask $s_p$. Accordingly, the noise $\boldsymbol{\epsilon}$ is cropped to $\boldsymbol{\epsilon}_p$. To avoid mismatches between the local patch/mask and user prompt, an empty prompt $\varnothing$ is used. These inputs are processed through the UNet to facilitate learning of local noise prediction. The loss function for micro-aware process is:
\begin {equation} 
 \begin{aligned} 
\mathcal{L}_{micro}=\|\boldsymbol{\epsilon}_{p}-f_{\theta}(\mathbf{z}_{t}^{(s_p)}, \mathbf{z}^{(x_p)},\boldsymbol{c}_{\varnothing}, t, \psi_b)\|_{2}^{2}.
\end{aligned}   
\end {equation}

\begin{figure}[t!]
\centering
\begin{overpic}[width=0.48\textwidth]{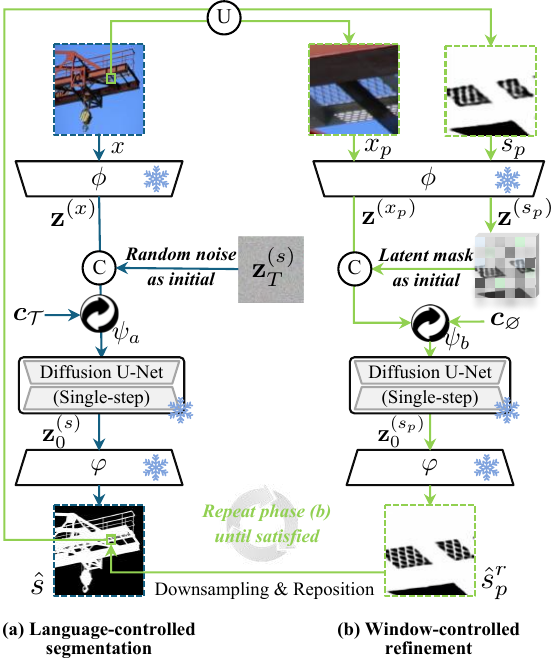}
\end{overpic}
\vspace*{-10pt}
\caption{Overview of the inference protocol, which consists of two steps. The first step is language-controlled segmentation, where \ourmodel~switches to the macro mode to generate an initial segmentation result based on the language prompt. If the user is not satisfied with the details, the second step, window-controlled refinement, is executed to refine details within a controllable window of variable resolution. The refined local patches are then used to replace their corresponding regions in the initial result. The second step can be repeated indefinitely until a satisfactory segmentation result is achieved.}
\label{fig:train_inference}
\end{figure}

\noindent\textbf{Joint training recipe.}
To achieve collaborative enhancement between the two modes, we jointly train the U-Net $f_{\theta}(\cdot)$ with both modes, as shown in Fig.~\ref{fig:train_test}. The naive diffusion loss Equ.\eqref{equ:naiveloss} is reformulated as the sum of the losses from each mode: $\mathcal{L}_{u} = \mathcal{L}_{macro}+\mathcal{L}_{micro}$. 

After training the U-Net, we fine-tune the VAE decoder $\varphi$ initially designed for RGB image reconstruction, to adapt it for the DIS task. First, to preserve the effectiveness of the previous training results, we freeze both the encoder $\phi$ and the U-Net during this process, while only fine-tuning the decoder $\varphi$. Second, we make simple structural adjustments to the decoder $\varphi$ by adding shortcut connections between the encoder and decoder. Moreover, the output layer of the decoder reduces the channels from 3 to 1, with its weight tensor initialized by averaging across the channels. Next, we randomly input pairs (image $x$, text prompt $\mathcal{T}$, $\psi_a$) or (local patch $x_p$, empty text prompt $\varnothing$, $\psi_b$) with a normally distributed noise $\mathbf{z}_{T}^{(s)} \in \mathcal{N}(\mathbf{0}, \mathbf{1})$ into the model. These input pass sequentially through the VAE encoder, U-Net, denoising scheduler, and VAE decoder to obtain the predicted segmentation mask $\hat{s}$ or local mask $\hat{s}_p$. The process is supervised using the annotated mask $s$ or $s_p$, with the structure loss function defined as follows:
\begin {equation} 
 \begin{aligned} 
\mathcal{L}_{d}=\begin{cases}
\mathcal{L}_{wbce}( \hat{s }, s )  + \mathcal{L}_{wiou}( \hat{s }, s )  & \text{ if } \psi=\psi_a , \\
\mathcal{L}_{wbce}( \hat{s}_p, s_p )  + \mathcal{L}_{wiou}( \hat{s}_p, s_p)  & \text{ if } \psi=\psi_b.
\end{cases}
\end{aligned}   
\end {equation}
where $\mathcal{L}_{wbce}(\cdot)$ and $\mathcal{L}_{wiou}(\cdot)$ are defined as \cite{yu2024multi,zhou2023dichotomous}. It is important to note that, in order to obtain $\hat{s}$ or  $\hat{s}_p$, the model needs to performs $T$ denoising steps to generate the clean latent features $\mathbf{z}_{0}^{(s)}$ or $\mathbf{z}_{0}^{(s_p)}$ from random noise. However, using denoising schedulers like DDIM \cite{song2020denoising}, which typically require 50 steps to generate segmentation images, proves impractical in realistic settings \cite{li2024controlnet++}. Therefore, we introduce the trajectory consistency distillation (TCD) \cite{zheng2024trajectory} as an out-of-the-box denoising scheduler to simplify the sampling process to a single step. It not only enables feasible fine-tuning of the VAE decoder $\varphi$ avoiding out-of-memory, but also enhances the inference efficiency.

\begin{table*}[ht]
\centering
\scriptsize
\renewcommand{\arraystretch}{1.2}
\renewcommand{\tabcolsep}{1.65mm}
\begin{tabular}{r|ccccccccccccccc}
& \multicolumn{5}{c|}{DIS-TE1 (500 images)} & \multicolumn{5}{c|}{DIS-TE2 (500 images)} & \multicolumn{5}{c}{DIS-TE3 (500 images)} \\ 
Methods &$F_\beta^\omega \uparrow$ &$F_\beta^{mx}\uparrow$ & $\mathcal{M} \downarrow$ & $\mathcal{S}_{\alpha} \uparrow$ & \multicolumn{1}{c|} {$E_\phi^{mn} \uparrow$}  & $F_\beta^\omega \uparrow$ &$F_\beta^{mx}\uparrow$  & $\mathcal{M} \downarrow$ & $\mathcal{S}_{\alpha} \uparrow$ & \multicolumn{1}{c|}{$E_\phi^{mn} \uparrow$}  & $F_\beta^\omega \uparrow$ &$F_\beta^{mx}\uparrow$  & $\mathcal{M} \downarrow$ & $\mathcal{S}_{\alpha} \uparrow$ & $E_\phi^{mn} \uparrow$  \\ \hline
BASNet$_{19}$ \cite{qin2019basnet} & 0.577 & 0.663 & 0.105 & 0.741 & \multicolumn{1}{c|}{0.756} & 0.653 & 0.738 & 0.096 & 0.781 & \multicolumn{1}{c|}{0.808}  & 0.714 &0.790 & 0.080 & 0.816 & 0.848  \\
U$^{2}$Net$_{20}$ \cite{qin2020u2} & 0.601 & 0.701 & 0.085 & 0.762 & \multicolumn{1}{c|}{0.783}  & 0.676 & 0.768 & 0.083 & 0.798 & \multicolumn{1}{c|}{0.825}  & 0.721 &0.813 & 0.073 & 0.823 & 0.856  \\
HRNet$_{20}$ \cite{wang2020deep} & 0.579 &0.668 & 0.088 & 0.742 & \multicolumn{1}{c|}{0.797} & 0.664 & 0.747 & 0.087 & 0.784 & \multicolumn{1}{c|}{0.840} & 0.700 & 0.784  & 0.080 & 0.805 & 0.869  \\
PGNet$_{22}$ \cite{xie2022pyramid} & 0.680 &0.754 & 0.067 & 0.800 & \multicolumn{1}{c|}{0.848} & 0.743 &0.807 & 0.065 & 0.833 & \multicolumn{1}{c|}{0.880} & 0.785 &0.843 & 0.056 & 0.844 & 0.911 \\
IS-Net$_{22}$ \cite{qin2022highly} & 0.662 &0.740 & 0.074 & 0.787 & \multicolumn{1}{c|}{0.820}  & 0.728 &0.799 & 0.070 & 0.823 & \multicolumn{1}{c|}{0.858}  & 0.758 & 0.830 & 0.064 & 0.836 & 0.883  \\
InSPyReNet$_{22}$ \cite{kim2022revisiting} & 0.788 &0.845 & 0.043 & 0.873 & \multicolumn{1}{c|}{0.894}  & 0.846 &0.894 & 0.036 & 0.905 & \multicolumn{1}{c|}{0.928} &  0.871 &0.919  & 0.034 & 0.918 & 0.943  \\
FP-DIS$_{23}$ \cite{zhou2023dichotomous} & 0.713 &0.784  & 0.060 & 0.821 & \multicolumn{1}{c|}{0.860}  & 0.767 &0.827 & 0.059 & 0.845 & \multicolumn{1}{c|}{0.893}  & 0.811 &0.868  & 0.049 & 0.871 & 0.922  \\
UDUN$_{23}$ \cite{pei2023unite} & 0.720 &0.784 & 0.059 & 0.817 & \multicolumn{1}{c|}{0.864}  & 0.768 &0.829 & 0.058 & 0.843 & \multicolumn{1}{c|}{0.886} & 0.809 &0.865 & 0.050 & 0.865 & 0.917 \\
BiRefNet$_{24}$ \cite{zheng2024birefnet} & 0.819 &0.860  & 0.037 & 0.885 & \multicolumn{1}{c|}{0.911}  & 0.857 &0.894 & 0.036 & 0.900 & \multicolumn{1}{c|}{0.930}  & 0.893 &0.925 & 0.028 & 0.919 & 0.955  \\
GenPercept$_{24}$ \cite{xu2024diffusion} & 0.794 &0.844 & 0.038 & 0.871 & \multicolumn{1}{c|}{0.909} & 0.828 &0.875 & 0.040 & 0.887 & \multicolumn{1}{c|}{0.925} &  0.840 &0.890 & 0.039 & 0.893 & 0.939  \\
MVANet$_{24}$ \cite{yu2024multi} & 0.820 &0.870 & 0.037 & 0.885 & \multicolumn{1}{c|}{0.914}  & 0.875 &0.915 & 0.030 & 0.917 & \multicolumn{1}{c|}{0.943} & 0.888 &0.929 & 0.029 & 0.923 & 0.953  \\
\mygray 
\textbf{Ours-S} & \secondcolor{ 0.886} & \secondcolor{ 0.917} & \secondcolor{ 0.025} & \bestcolor{ \textbf{0.917}} & \multicolumn{1}{c|}{\secondcolor{ 0.946}}  & \secondcolor{ 0.906} & \secondcolor{ 0.934} & \secondcolor{ 0.024} & \bestcolor{ \textbf{0.932}} & \multicolumn{1}{c|}{\secondcolor{ 0.958}}  & \secondcolor{ 0.908} & \secondcolor{ 0.937} & \secondcolor{ 0.025} & \secondcolor{ 0.931} & \secondcolor{ 0.960}   \\
\mygray 
\textbf{Ours-R} & \bestcolor{ \textbf{0.890}} & \bestcolor{ \textbf{0.919}} & \bestcolor{ \textbf{0.024}} & \bestcolor{ \textbf{0.917}} & \multicolumn{1}{c|}{\bestcolor{ \textbf{0.948}}}  & \bestcolor{ \textbf{0.914}} & \bestcolor{ \textbf{0.936}} & \bestcolor{ \textbf{0.022}} & \bestcolor{ \textbf{0.932}} & \multicolumn{1}{c|}{\bestcolor{ \textbf{0.961}}}  & \bestcolor{ \textbf{0.919}} & \bestcolor{ \textbf{0.942}} & \bestcolor{ \textbf{0.022}} & \bestcolor{ \textbf{0.932}} & \bestcolor{ \textbf{0.964}}  \\ \hline
& \multicolumn{5}{c|}{DIS-TE4 (500 images)} & \multicolumn{5}{c|}{DIS-TE (1-4) (2,000 images)} & \multicolumn{5}{c}{DIS-VD (470 images)} \\ 
Methods &$F_\beta^\omega \uparrow$ &$F_\beta^{mx}\uparrow$ & $\mathcal{M} \downarrow$ & $\mathcal{S}_{\alpha} \uparrow$ & \multicolumn{1}{c|}{$E_\phi^{mn} \uparrow$}  & $F_\beta^\omega \uparrow$ &$F_\beta^{mx}\uparrow$  & $\mathcal{M} \downarrow$ & $\mathcal{S}_{\alpha} \uparrow$ & \multicolumn{1}{c|}{$E_\phi^{mn} \uparrow$}  & $F_\beta^\omega \uparrow$ &$F_\beta^{mx}\uparrow$  & $\mathcal{M} \downarrow$ & $\mathcal{S}_{\alpha} \uparrow$ & $E_\phi^{mn} \uparrow$ \\ \hline
BASNet$_{19}$ \cite{qin2019basnet} & 0.713 & 0.785 & 0.087 & 0.806 & \multicolumn{1}{c|}{0.844}  & 0.664 & 0.744 & 0.092 & 0.786 & \multicolumn{1}{c|}{0.814}  & 0.656 & 0.737 & 0.094 & 0.781 & 0.809  \\
U$^{2}$Net$_{20}$ \cite{qin2020u2} & 0.707 & 0.800 & 0.085 & 0.814 & \multicolumn{1}{c|}{0.837}  & 0.676 & 0.771 & 0.082 & 0.799 & \multicolumn{1}{c|}{0.825}  & 0.656 & 0.753 & 0.089 & 0.785 & 0.809  \\
HRNet$_{20}$ \cite{wang2020deep} & 0.687 & 0.772 & 0.092 & 0.792 & \multicolumn{1}{c|}{0.854} & 0.658 & 0.743 & 0.087 & 0.781 & \multicolumn{1}{c|}{0.840}  & 0.641 & 0.726 & 0.095 & 0.767 & 0.824  \\
PGNet$_{22}$ \cite{xie2022pyramid} & 0.774 & 0.831 & 0.065 & 0.841 & \multicolumn{1}{c|}{0.899}& 0.746 & 0.809 & 0.063 & 0.830 & \multicolumn{1}{c|}{0.885}  & 0.733 & 0.798 & 0.067 & 0.824 & 0.879  \\
IS-Net$_{22}$ \cite{qin2022highly} & 0.753 & 0.827 & 0.072 & 0.830 & \multicolumn{1}{c|}{0.870}  & 0.726 & 0.799 & 0.070 & 0.819 & \multicolumn{1}{c|}{0.858}  & 0.717 & 0.791 & 0.074 & 0.813 & 0.856  \\
InSPyReNet$_{22}$ \cite{kim2022revisiting} & 0.848 & 0.905 & 0.042 & 0.905 & \multicolumn{1}{c|}{0.928}  & 0.838 & 0.891 & 0.039 & 0.900 & \multicolumn{1}{c|}{0.923}  & 0.834 & 0.889 & 0.042 & 0.900 & 0.922  \\
FP-DIS$_{23}$ \cite{zhou2023dichotomous} & 0.788 & 0.846 & 0.061 & 0.852 & \multicolumn{1}{c|}{0.906}  & 0.770 & 0.831 & 0.047 & 0.847 & \multicolumn{1}{c|}{0.895}  & 0.763 & 0.823 & 0.062 & 0.843 & 0.891  \\
UDUN$_{23}$ \cite{pei2023unite} & 0.792 & 0.846 & 0.059 & 0.849 & \multicolumn{1}{c|}{0.901}  & 0.772 & 0.831 & 0.057 & 0.844 & \multicolumn{1}{c|}{0.892}  & 0.763 & 0.823 & 0.059 & 0.838 & 0.892  \\
BiRefNet$_{24}$ \cite{zheng2024birefnet} & 0.864 & 0.904 & 0.039 & 0.900 & \multicolumn{1}{c|}{0.939}  & 0.858 & 0.896 & 0.035 & 0.901 & \multicolumn{1}{c|}{0.934}  & 0.854 & 0.891 & 0.038 & 0.898 & 0.931  \\
GenPercept$_{24}$ \cite{xu2024diffusion} & 0.801 & 0.861 & 0.055 & 0.869 & \multicolumn{1}{c|}{0.918}  & 0.816 & 0.868 & 0.043 & 0.880 & \multicolumn{1}{c|}{0.923}  & 0.815 & 0.865 & 0.043 & 0.881 & 0.922  \\
MVANet$_{24}$ \cite{yu2024multi} & 0.866 & 0.913 & 0.038 & 0.910 & \multicolumn{1}{c|}{0.940}  & 0.862 & 0.907 & 0.034 & 0.909 & \multicolumn{1}{c|}{0.938}  & 0.861 & 0.904 & 0.035 & 0.909 & 0.937  \\
\mygray 
\textbf{Ours-S} & \secondcolor{ 0.890} & \secondcolor{ 0.926} & \secondcolor{ 0.032} & \secondcolor{ 0.920} & \multicolumn{1}{c|}{\secondcolor{ 0.955}}  & \secondcolor{ 0.898} & \secondcolor{ 0.929} & \secondcolor{ 0.027} & \secondcolor{ 0.925} & \multicolumn{1}{c|}{\secondcolor{ 0.955}}  & \secondcolor{ 0.894} & \secondcolor{ 0.925} & \secondcolor{ 0.026} & \secondcolor{ 0.924} & \secondcolor{ 0.955}  \\
\mygray 
\textbf{Ours-R} & \bestcolor{ \textbf{0.910}} & \bestcolor{ \textbf{0.932}} & \bestcolor{ \textbf{0.026}} & \bestcolor{ \textbf{0.922}} & \multicolumn{1}{c|}{\bestcolor{ \textbf{0.964}}}  & \bestcolor{ \textbf{0.908}} & \bestcolor{ \textbf{0.932}} & \bestcolor{ \textbf{0.024}} & \bestcolor{ \textbf{0.926}} & \multicolumn{1}{c|}{\bestcolor{ \textbf{0.959}}}  & \bestcolor{ \textbf{0.905}} & \bestcolor{ \textbf{0.929}} & \bestcolor{ \textbf{0.023}} & \bestcolor{ \textbf{0.925}} & \bestcolor{ \textbf{0.959}}  \\  
\end{tabular}
\vspace*{-8pt}
\caption{Quantitative comparison of DIS5K with 11 representative methods. $\downarrow$ represents the lower value is better, while $\uparrow$ represents the higher value is better. The best and the second-best results are highlighted in \bestcolor{\textbf{red}} and \secondcolor{blue}.}
\label{tab:Quantitative_woHCE}
\end{table*}

\subsection{Two-stage Inference}
\label{sec:Local Refine}
Fig.~\ref{fig:train_inference} provides an overview of the inference process, which consists of two steps. The first step is language-guided segmentation, where \ourmodel~switches to macro mode to generate segmentation results based on language prompts. The second step serves as an optional refinement stage, invoked only when user-driven adjustments are required. During this process, the micro mode is selected, allowing the user to refine details within a controllable window of variable resolution. It can be repeated indefinitely until a satisfactory segmentation result is achieved. The details of both steps are provided below.
 
\noindent\textbf{Language-controlled segmentation.}~The switcher is modulated by $\psi_a$ which activates the macro mode. Given the input image $x$, we use the VAE encoder $\phi$ to transform it into the latent space $\mathbf{z}^{(x)}=\phi(x)$. Then, we initialize the segmentation mask latent as standard Gaussian noise $\mathbf{z}_{T}^{(s)} \in \mathcal{N}(\mathbf{0}, \mathbf{1})$. The language prompts used to control the segmented objects $\mathcal{T}$ can be generated by VLM or customized by users. Next, they are fed into the denoising U-Net together to predict noise. A single-step denoising is performed with the TCD scheduler \cite{zheng2024trajectory} to obtain the clean latent features $\mathbf{z}_{0}^{(s)}$. Finally, by decoding the features $\mathbf{z}_{0}^{(s)}$ with the fine-tuned VAE decoder, we obtain the language-controlled segmentation map $\hat{s }= \varphi(\mathbf{z}_{0}^{(s)})$.

\noindent\textbf{Window-controlled refinement.}~The switcher is set to $\psi_b$ which activates the micro mode. A key challenge is ensuring reliable refinement when feeding local patches cropped from arbitrary windows into the network, as they lack context from the full image. To address this, we propose using local patches from the global segmentation result instead of noise as the starting point for diffusion, indirectly transferring contextual information between the two modes.

Specifically, the user can click on any area in the initial segmentation map to select the unsatisfactory region as the window for refinement. For each window for refinement, two patches are cropped: the local patch $x_p$, cropped from the input image $x$, and the initial local mask $\hat{s}_p$, cropped from the language-controlled segmentation map $\hat{s}$ obtained in the first step. These patches are upsampled to the input size of the model. Next, we utilize the encoder $\phi$ to obtain the latent features of the local patch $\mathbf{z}^{(x_p)}=\phi(x_p)$ and the local mask $\mathbf{z}^{(\hat{s}_p)}=\phi(\hat{s}_p)$, respectively. We change the language prompt $\mathcal{T}$ to empty $\varnothing$ as a condition. The latent features of the initial local mask $\mathbf{z}^{(\hat{s}_p)}$ are combined with those of the local patch $\mathbf{z}^{(x_p)}$ and fed into the denoising U-Net to predict noise. After denoising with the TCD denoising scheduler and decoding with the VAE decoder, a refined local mask $\hat{s}^r_p$ with clearer details is obtained. Finally, $\hat{s}^r_p$ replaces the original content at its designated location in the language-controlled segmentation map $\hat{s}$. This process allows multiple windows to be refined simultaneously. When there is an overlapping region between multiple windows, the value from the window with a larger proportion of the overlapping area is chosen as the final optimization result.

\vspace{-3pt}
\section{Experiments}
\label{sec:experiments}
\subsection{Experimental setups}
\noindent\textbf{Dataset.}~We perform all experiments on the DIS5K benchmark \cite{qin2022highly}, which contains 5,470 high-resolution image-mask pairs across 225 semantic categories. This benchmark is divided into DIS-TR (3,000 images), DIS-VD (470 images), and DIS-TE  (2,000 images). We conduct all training on DIS-TR and evaluate all models on DIS-VD and DIS-TE. The DIS-TE has four subsets, from DIS-TE1 to DIS-TE4, each containing 500 samples, to represent increasing levels of shape complexities.

\noindent\textbf{Evaluation.}~For a comprehensive comparison, we employ five widely used pixel-wise metrics to assess the ability of the models, including the weighted F-measure ($F_\beta^\omega$) \cite{6909433}, the maximum F-measure ($F_\beta^{mx}$) \cite{perazzi2012saliency}, the structure measure ($\mathcal{S}_{\alpha}$) \cite{cheng2021structure}, the mean enhanced-alignment measure ($E_\phi^{mn}$) \cite{fan2021cognitive}, and the mean absolute error ($M$) \cite{perazzi2012saliency}. 

\noindent\textbf{Implementation details.}~We implement \ourmodel~using PyTorch library, accelerated by a single NVIDIA A100-40GB GPU. To repurpose Stable Diffusion v2 \cite{rombach2022high} for conditional mask generation, we duplicate the diffusion U-Net's input layer to match the concatenated image feature and noised mask feature. We initialise the duplicated layers via copying the weight tensor and halve its values to avoid inflation of activation size \cite{ke2024repurposing}. We employ the DDPM noise scheduler \cite{ho2020denoising} with 1000 steps in the diffusion U-Net training stage, while the VAE decoder is trained using a single-step TCD scheduler \cite{zheng2024trajectory} due to computational constraints. Both stages use a batch size of 32 and Adam optimizer with a 3e-5 learning rate, while diffusion U-Net is fine-tuned for 30K iterations and VAE decoder for 6K iterations. To increase visual diversity, a random horizontal flipped augmentation and an annealed multi-resolution noise strategy are applied. For better inference efficiency, we use the TCD noise scheduler in both macro and micro modes. All inputs, whether full-size images or window-sized regions, are uniformly resized to $1024^2$px for training and inference.

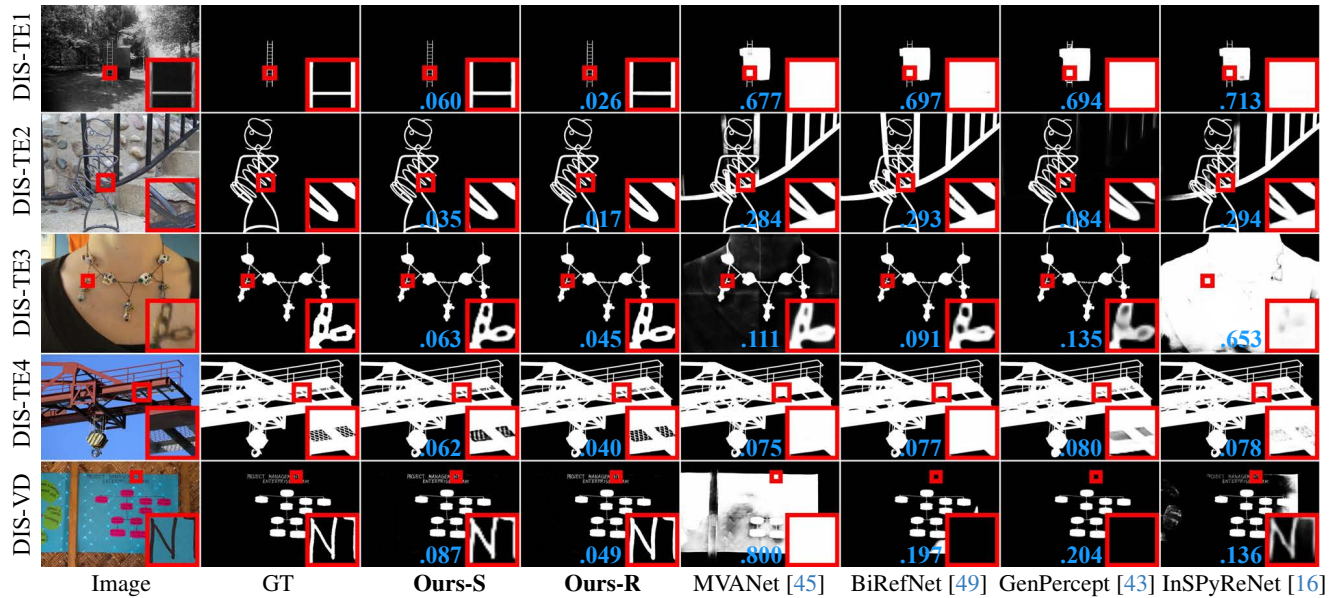
\begin{figure*}[t!]
\centering
\begin{overpic}[width=\textwidth]{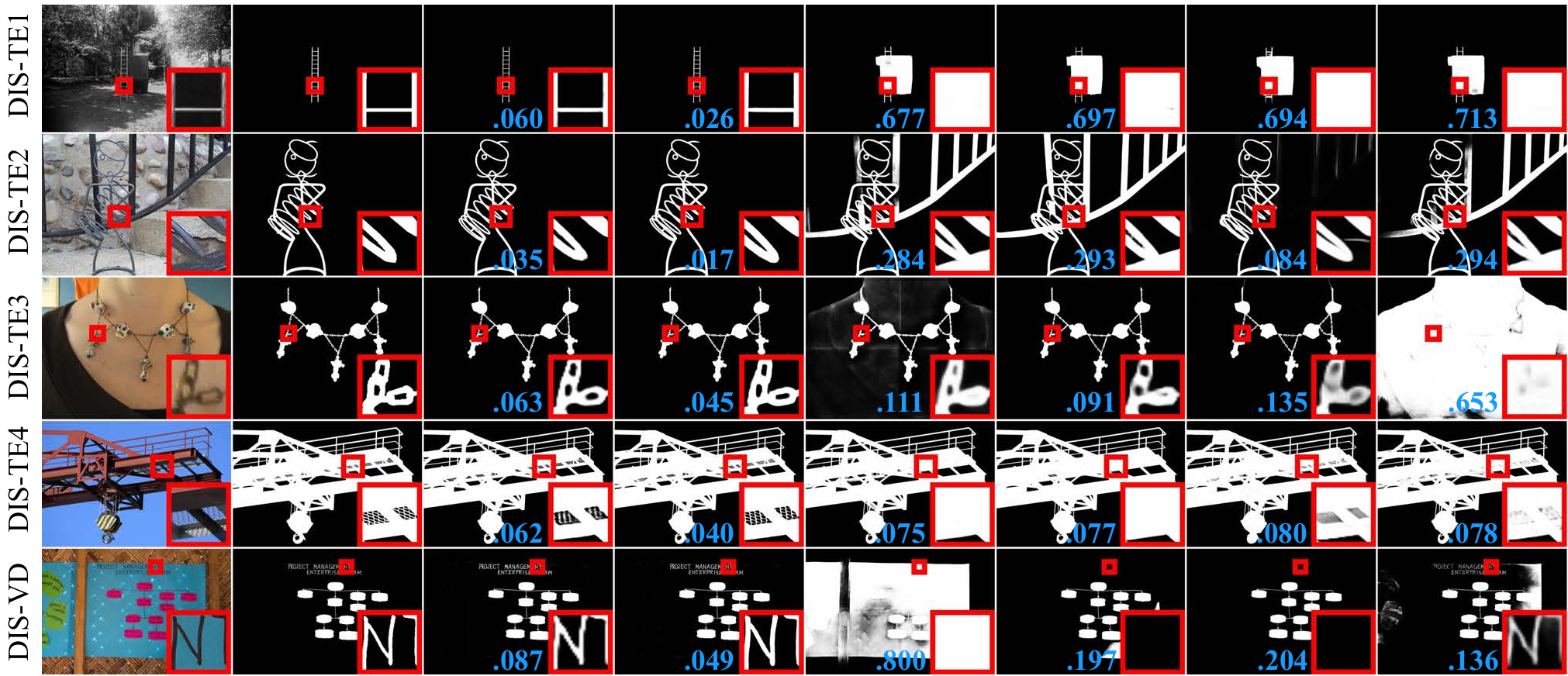}
    \put(6.5,-1.8){\small Image}
    \put(19.5,-1.8){\small GT}
    \put(31,-1.8){\small \textbf{Ours-S}}
    \put(42.5,-1.8){\small \textbf{Ours-R}}
    \put(52.3,-1.8){\small MVANet~\cite{yu2024multi}}
    \put(64.3,-1.8){\small BiRefNet~\cite{zheng2024birefnet}}
    \put(75.6,-1.8){\small GenPercept~\cite{xu2024diffusion}}
    \put(88,-1.8){\small InSPyReNet~\cite{kim2022revisiting}}

\end{overpic}
\vspace*{-8pt}
\caption{
Qualitative comparison of our model with four leading models. Local masks are evaluated with $\mathcal{M}$ score for clarity.
}
\vspace*{-8pt}
\label{fig:Qualitative_comparisons1}
\end{figure*}

\subsection{Comparison with SOTA Methods}
\noindent\textbf{Quantitative comparison.}~As presented in Tab.~\ref{tab:Quantitative_woHCE}, we establish a DIS benchmark with 11 well-known task-related methods, including BASNet \cite{qin2019basnet}, U$^{2}$Net \cite{qin2020u2}, HRNet \cite{wang2020deep}, PGNet \cite{xie2022pyramid}, IS-Net \cite{qin2022highly}, InSPyReNet \cite{kim2022revisiting}, FP-DIS \cite{zhou2023dichotomous}, UDUN \cite{pei2023unite}, BiRefNet \cite{zheng2024birefnet}, GenPercept \cite{xu2024diffusion}, and MVANet \cite{yu2024multi}. With the same setup of input size ($1024^2$px), our basic configuration with language controls (Ours-S), has outperformed the competing methods \cite{wang2020deep,xie2022pyramid,kim2022revisiting,zhou2023dichotomous,pei2023unite,xu2024diffusion,yu2024multi,zheng2024birefnet,qin2022highly} in all metrics across all test sets. These results underscore the efficacy of \ourmodel~with user prompts in addressing the challenging cases of DIS, especially for diverse categories of targets. For example, Ours-S surpasses the runner-up model MVANet \cite{yu2024multi} by 6.6\% in $F_\beta^\omega$ on DIS-TE1.
In micro mode, Ours-R refines the initial predictions from the macro mode (Ours-S), leading to a further 2.0\% increase in $F_\beta^\omega$ on DIS-TE4. Here, to simulate subjectivity of user interaction, window candidates are chosen along the object's contours of the GT. See \hyperlink{supplementary_window_anchor}{\textcolor{magenta}{supp.}} for more details. By integrating dual control mechanisms into \ourmodel, we achieve a gain of 7.0\% in $F_\beta^\omega$ on DIS-TE1 compared to MVANet.

\noindent\textbf{Qualitative comparison.} Fig.~\ref{fig:Qualitative_comparisons1} provides a qualitative comparison between our approach and the most competitive existing DIS models. At the macro level, we achieve more complete segmentation of the target regions, while at the micro level, it demonstrates superior precision in handling complex structures and intricate details. For example, as shown in the 5th row, the edges of the subtitles appear sharper, and the $\mathcal{M}$ score for local patches shows a drop of 3.8\% from Ours-S (0.087) to Ours-R (0.049).

\subsection{Diagnostic Study}
Next, we assess the impact of key components through a series of ablation studies conducted on DIS-TE4 subset.

\begin{table}[t!]
\centering
\scriptsize
\renewcommand{\arraystretch}{1}
\renewcommand{\tabcolsep}{2.7mm}
\begin{tabular}{r|llll}
Ablation settings & $F_\beta^{mx}\uparrow$ & $\mathcal{M} \downarrow$ & $\mathcal{S}_{\alpha} \uparrow$ & $E_\phi^{mn} \uparrow$ \\ \hline
baseline \cite{rombach2022high} & 0.904  & 0.047 & 0.904 & 0.916 \\
\textit{w/o} micro-level training & 0.912  & 0.037 & 0.909 & 0.943 \\
\textit{w/o} fine-tuning VAE decoder &  0.919  & 0.040  & 0.915  &  0.933 \\
\hline \mygray 
\textbf{Ours-S} & {\textbf{0.926}}  &  \textbf{0.032} & {\textbf{0.920}} & {\textbf{0.955}} \\ 
\end{tabular}
\vspace*{-5pt}
\caption{Ablation study for the baseline and general settings.}
\label{tab:ablation1}
\vspace*{-5pt}
\end{table}

\noindent\textbf{Baseline and general settings.}~To reveal the true potential of stable diffusion \cite{rombach2022high} in the DIS task, we initialise a baseline model by modifying \ourmodel~with three changes: omitting the mode switcher, training the diffusion U-Net without user prompts, and not fine-tuning VAE decoder. As presented in the first row of Tab.~\ref{tab:ablation1}, the baseline variant does not achieve superior performance. To assess the impact of the joint training strategy, we remove the mode switcher and train \ourmodel~exclusively using language controls. This variant (second row of Tab.~\ref{tab:ablation1}) exhibits performance drops across all metrics relative to Ours-S, demonstrating the dual-mode synergy's role in providing scalable geometric representations of for varied input sizes. In addition, we create another variant (third row of Tab.~\ref{tab:ablation1}) where the VAE decoder is not fine-tuned, and its output is averaged across channels to produce a single-channel mask prediction. The inferior performance demonstrates that fine-tuning the VAE decoder is essential for high-resolution segmentation, as it enables the decoding process to complement the denoised mask features with fine-grained details.

\begin{table}[t!]
\centering
\scriptsize
\renewcommand{\arraystretch}{1}
\renewcommand{\tabcolsep}{2.5mm}
\begin{tabular}{r|llll}
Ablation settings & $F_\beta^{mx}\uparrow$ & $\mathcal{M} \downarrow$ & $\mathcal{S}_{\alpha} \uparrow$ & $E_\phi^{mn} \uparrow$ \\ \hline
\textit{w/o} user prompt (train \& test) & 0.912  & 0.036 & 0.908 & 0.944 \\
\textit{w/o} user prompt (test only) & 0.915  & 0.036 & 0.909 & 0.947 \\
\hline \mygray 
\textbf{\textit{w/} user prompt (train \& test)} & {\textbf{0.926}}  &  \textbf{0.032} & {\textbf{0.920}} & {\textbf{0.955}} \\ 
\end{tabular}
\vspace*{-5pt}
\caption{Ablation study on macro controls for LawDIS.}
\label{tab:ablation}
\vspace*{-10pt}
\end{table}

\begin{figure}[t!]
\centering
\begin{overpic}[width=0.48\textwidth]{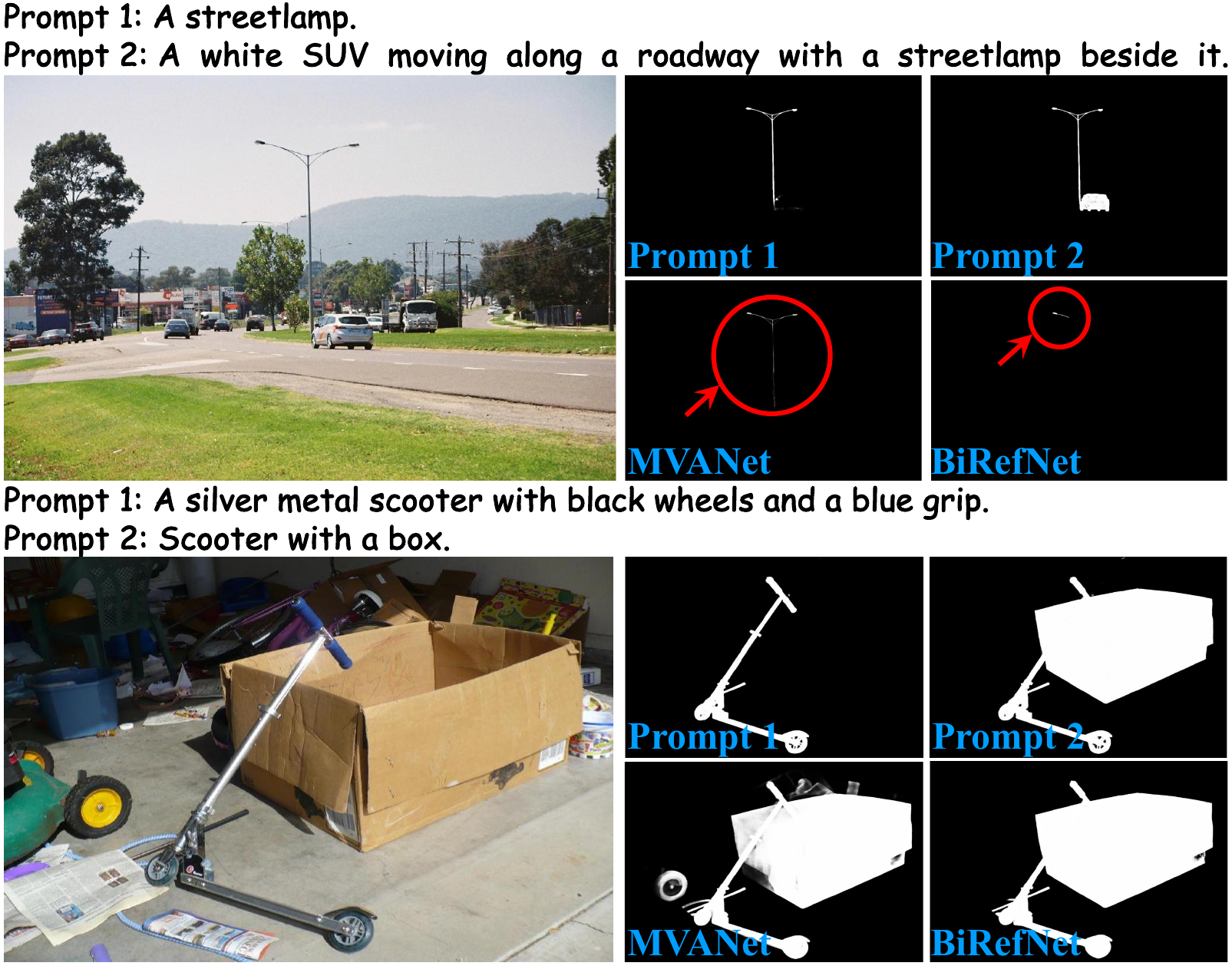}
\end{overpic}
\vspace*{-20pt}
\caption{Qualitative predictions under different macro controls.}
\vspace*{-10pt}
\label{fig:prompt_compare}
\end{figure}

\noindent\textbf{Effectiveness of macro controls.}
We design experiments under two settings: in the first row of Tab.~\ref{tab:ablation}, the user prompts are set to empty during the training and testing phases; in the second row, user prompts are omitted only during the testing phase. 
Compared to them, our default setting, which uses user prompts in both phases, achieves superior results. As shown in Fig.~\ref{fig:prompt_compare}, our model showcases the ability to flexibly segment various target objects based on customized language prompts. In contrast, other methods \cite{yu2024multi,zheng2024birefnet}, which lack the ability to process language prompts, yield only fixed results through memory patterns during training, further emphasizing the effectiveness of \ourmodel.

\begin{table}[!t]
\centering
\scriptsize
\renewcommand{\arraystretch}{1}
\renewcommand{\tabcolsep}{1.85mm}
\begin{tabular}{r|cccc}
Ablation settings & $F_\beta^\omega \uparrow$ & $\mathcal{M} \downarrow$  & $BIoU^m \uparrow$  & $HCE_\gamma \downarrow$\\ 
\hline
basic setting (\ie, Ours-S) & 0.890 & 0.032  & 0.795 & 2481 \\
\hline
init. from Gaussian noise & -4.7\% & +1.9\%  & -7.1\% & -863 \\
auto windows selection & +1.7\% & -0.5\%  & +2.9\% & -767 \\
\hline \mygray 
\textbf{semi-auto windows selection} & \textbf{+2.0\%} & \textbf{-0.6\%}  & \textbf{+3.2\%} & \textbf{-871} \\ 
\end{tabular}
\vspace*{-5pt}
\caption{Ablation study on micro controls for \ourmodel.}
\label{tab:wr_ablation}
\vspace*{-15pt}
\end{table}

\noindent\textbf{Effectiveness of micro controls.}
To better unveil performance change on fine structures, we introduce two extra metrics, human correction effort ($HCE_\gamma$) \cite{qin2022highly} and boundary intersection-over-union ($BIoU^m$) \cite{cheng2021boundary}. In the second row of Tab.~\ref{tab:wr_ablation}, we replace the patch mask latent with Gaussian noise as input, resulting in a performance drop from 0.890 to 0.843 in $F_\beta^\omega$. This suggests that initiating the diffusion process using the segmentation results can help the model achieve more refined masks. In addition, the third row of Tab.~\ref{tab:wr_ablation} presents a fully automated window selection process, where windows are chosen around object edges in the initial predictions from Ours-S without user intervention. The result demonstrates that the WR strategy effectively improves segmentation performance, even without user intervention in the window selection.

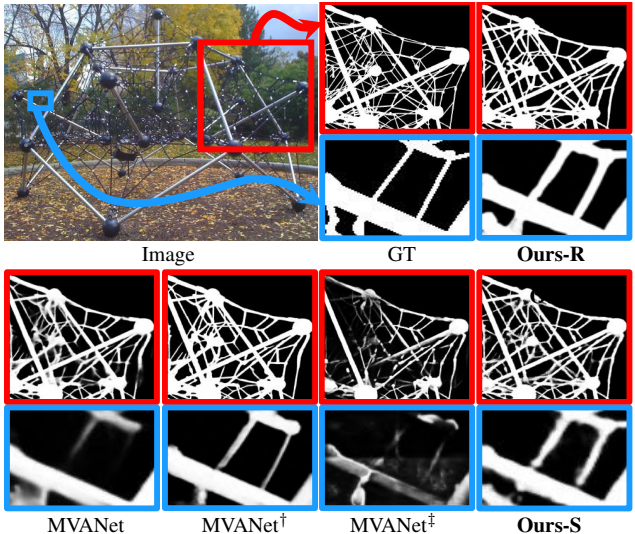
\begin{figure}[t!]
\centering
\begin{overpic}[width=0.48\textwidth]{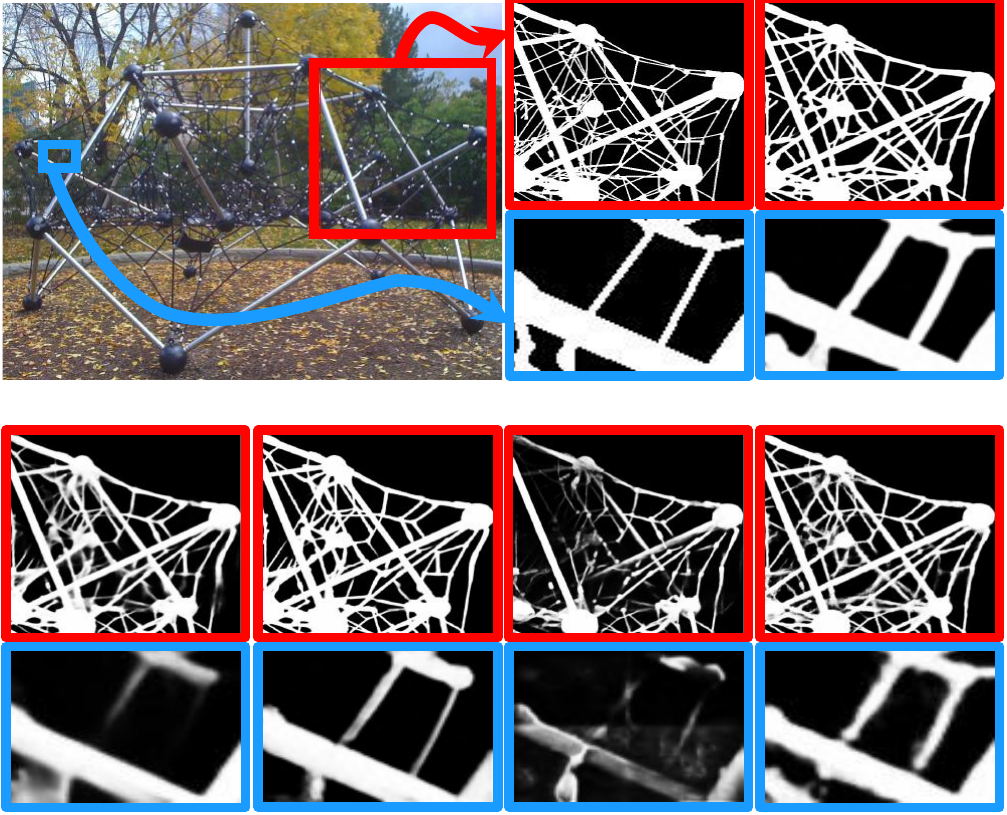}
\put(22,40){\footnotesize Image}
\put(61,40){\footnotesize GT}
\put(81.5,40){\footnotesize \textbf{Ours-R}}
\put(83.5,33.3){\footnotesize \textbf{Ours-S}}
\put(7,-3){\footnotesize MVANet}
\put(31.5,-3){\footnotesize MVANet$^\dagger$}
\put(55,-3){\footnotesize MVANet$^\ddagger$}
\put(81.5,-3){\footnotesize \textbf{Ours-S}}
\end{overpic}
\vspace*{-5pt}
\caption{Qualitative comparison of refinement performance of WR strategy and segmentation performance on local patches of MVANet \cite{yu2024multi}. $\dagger$ represents the refined segmentation results by applying WR strategy, while $\ddagger$ represents the segmentation result of the local patch obtained by other methods (\eg, MVANet).}
\vspace*{-10pt}
\label{fig:local_compare}
\end{figure}

\subsection{Discussion}

\begin{table}[!t]
\centering
\scriptsize
\renewcommand{\arraystretch}{1}
\renewcommand{\tabcolsep}{1.57mm}
\begin{tabular}{r|cccc}
 Methods & $F_\beta^\omega \uparrow$ & $\mathcal{M} \downarrow$ & $\mathcal{S}_{\alpha} \uparrow$ & $E_\phi^{mn} \uparrow$    \\ \hline
IS-Net$^\dagger$ \cite{qin2022highly} & 0.753 \textcolor{gray}{\underline{+6.4\%}}  & 0.072 \textcolor{gray}{\underline{-1.6\%}} & 0.830 \textcolor{gray}{\underline{+2.7\%}} & 0.870  \textcolor{gray}{\underline{+4.2\%}} \\
InSPyReNet$^\dagger$ \cite{kim2022revisiting} & 0.848 \textcolor{gray}{\underline{+4.2\%}}  & 0.042 \textcolor{gray}{\underline{-1.1\%}} & 0.905 \textcolor{gray}{\underline{+0.8\%}} & 0.928 \textcolor{gray}{\underline{+2.6\%}} \\
UDUN$^\dagger$ \cite{pei2023unite} & 0.792 \textcolor{gray}{\underline{+3.9\%}}  & 0.059 \textcolor{gray}{\underline{-1.0\%}} & 0.849 \textcolor{gray}{\underline{+2.0\%}} & 0.901 \textcolor{gray}{\underline{+2.2\%}} \\
BiRefNet$^\dagger$ \cite{zheng2024birefnet} & 0.864 \textcolor{gray}{\underline{+2.5\%}}  & 0.039 \textcolor{gray}{\underline{-0.6\%}} & 0.900 \textcolor{gray}{\underline{+0.7\%}} & 0.939 \textcolor{gray}{\underline{+1.3\%}} \\
MVANet$^\dagger$ \cite{yu2024multi} & 0.866 \textcolor{gray}{\underline{+3.2\%}}  & 0.038 \textcolor{gray}{\underline{-0.9\%}} & 0.910 \textcolor{gray}{\underline{+0.6\%}} & 0.940 \textcolor{gray}{\underline{+1.8\%}}\\ 
\end{tabular}
\vspace*{-5pt}
\caption{Using WR strategy as a post-refinement tool to enhance current DIS methods. The performance gains are underlined.}
\label{tab:wr_refine_other_methods}
\vspace*{-10pt}
\end{table}

\noindent\textbf{Can the proposed WR strategy function as a general post-refinement tool?} To reveal its compatibility, we apply the WR strategy to refine the initial masks predicted by various out-of-the-box DIS methods \cite{qin2022highly,pei2023unite,kim2022revisiting,zheng2024birefnet,yu2024multi}. As indicated in Tab.~\ref{tab:wr_refine_other_methods}, we improve the performance of each model in the DIS-TE4 subset to varying degrees. For example, the predictions of IS-Net \cite{qin2022highly} and MVANet \cite{yu2024multi} achieve 6.4\% and 3.2\% increases in $F_\beta^\omega$, respectively. Fig.~\ref{fig:local_compare} compares the zoom-in region from the full-image segmentation strategy (MVANet) to the refined patch mask obtained via our WR strategy (MVANet$^\dagger$). This demonstrate the feasibility of WR strategy as a post-refinement tool to enhance existing DIS methods' accuracy. Since stable diffusion generates masks by denoising from noise, modify the starting point using the latent features of an existing segmentation result patch instead, enabling the model to focus on refining details rather than rediscovering the overall structure.

\begin{table}[!t]
\centering
\scriptsize
\renewcommand{\arraystretch}{1}
\renewcommand{\tabcolsep}{1.15mm}
\begin{tabular}{r|cccc}
\multicolumn{1}{r|}{Methods} &
  $F_\beta^\omega \uparrow$ &
  $\mathcal{M} \downarrow$ &
  $\mathcal{S}_{\alpha} \uparrow$ &
  $E_\phi^{mn} \uparrow$ \\ \hline
IS-Net$^\ddagger$ \cite{qin2022highly} &  0.753 \textcolor{gray}{\underline{-3.6\%}} & 0.072 \textcolor{gray}{\underline{+1.0\%}} & 0.830 \textcolor{gray}{\underline{-4.5\%}} & 0.870 \textcolor{gray}{\underline{-4.5\%}} \\
InSPyReNet$^\ddagger$ \cite{kim2022revisiting} &  0.848 \textcolor{gray}{\underline{-0.8\%}} & 0.042 \textcolor{gray}{\underline{+0.7\%}} & 0.905 \textcolor{gray}{\underline{-2.0\%}}& 0.928 \textcolor{gray}{\underline{-0.5\%}} \\
UDUN$^\ddagger$ \cite{pei2023unite} & 0.792 \textcolor{gray}{\underline{-3.8\%}} & 0.059 \textcolor{gray}{\underline{+1.6\%}} & 0.849 \textcolor{gray}{\underline{-3.6\%}} & 0.901 \textcolor{gray}{\underline{-2.8\%}}  \\
BiRefNet$^\ddagger$ \cite{zheng2024birefnet} & 0.864 \textcolor{gray}{\underline{-3.9\%}} & 0.039 \textcolor{gray}{\underline{+1.6\%}} & 0.900 \textcolor{gray}{\underline{-4.3\%}} & 0.939 \textcolor{gray}{\underline{-3.1\%}}  \\
MVANet$^\ddagger$ \cite{yu2024multi} & 0.866 \textcolor{gray}{\underline{-0.2\%}} & 0.038 \textcolor{gray}{\underline{+0.6\%}} & 0.910 \textcolor{gray}{\underline{-1.4\%}} & 0.940 \textcolor{gray}{\underline{-0.7\%}}  \\ \hline 
\mygray 
\textbf{Ours$^\ddagger$ (Ours-R)} & 0.890 \textbf{+ 2.0\%} & 0.032 \textbf{- 0.6\%} & 0.920 \textbf{+ 0.2\%} & 0.955 \textbf{+ 0.9\%} \\ 

\end{tabular}
\vspace*{-5pt}
\caption{Quantitative comparison of the segmentation performance on local patches.}
\label{tab:local_compare}
\vspace*{-10pt}
\end{table}

\noindent\textbf{Are existing DIS methods compatible with local patch inputs?} To investigate this, we adopt the same window selection approach and patch replacement technique as default in \ourmodel, and feed local patches\footnote{Patches are resized to each model's required input resolution.} to each model to obtain masks. As illustrated in Tab.~\ref{tab:local_compare}, except for ours, all earlier methods exhibit varying levels of performance decline. For example, the $F_\beta^\omega$ of UDUN \cite{pei2023unite} and BiRefNet \cite{zheng2024birefnet} decreases by 3.8\% and 3.9\%, respectively. Additionally, Fig.~\ref{fig:local_compare} demonstrates the comparison of MVANet's segmentation on the full image (MVANet) versus local patches (MVANet$^\ddagger$), showing a notable performance drop for the latter setting. This indicates earlier methods relied on inputs of fixed resolutions do not accommodate variable inputs.

\begin{table}[!t]
\centering
\scriptsize
\renewcommand{\arraystretch}{1}
\renewcommand{\tabcolsep}{1mm}
\begin{tabular}{r|c|c|c|cc|c}
Methods & Fine-tuning VAE & Scheduler & Infer step & $F_\beta^\omega \uparrow$ & $\mathcal{M} \downarrow$ & FPS $\uparrow$ \\ 
\hline
MVANet \cite{yu2024multi} & - & - & - & 0.866 & 0.038 & 1.40 \\ 
\hline
\multirow{5}{*}{Ours-S} 
 & $\times$ &  DDPM \cite{ho2020denoising}  & 500  & 0.867 & 0.038 & 0.006 \\
 & $\times$ & DDIM \cite{song2020denoising} & 10 & 0.835  & 0.044  & 0.55 \\
 & $\times$ & DDIM \cite{song2020denoising} & 50 & 0.842  & 0.043  & 0.12 \\
 & $\times$ & TCD \cite{zheng2024trajectory}  & 1  & 0.856  & 0.040  & \textbf{3.09} \\
 & \cellcolor[HTML]{f5f5f5}\checkmark
 & \cellcolor[HTML]{f5f5f5}TCD \cite{zheng2024trajectory}
 & \cellcolor[HTML]{f5f5f5}1
 & \cellcolor[HTML]{f5f5f5}\textbf{0.890}
 & \cellcolor[HTML]{f5f5f5}\textbf{0.032}
 & \cellcolor[HTML]{f5f5f5}3.07 \\
\end{tabular}
\vspace*{-5pt}
\caption{Efficiency analysis of~\ourmodel.}
\label{tab:efficiency}
\vspace*{-10pt}
\end{table}

\noindent\textbf{Efficiency analysis.} We conduct an efficiency analysis using a single A100 GPU. First, we evaluate the results of denoising with the original DDPM scheduler for 500 steps, as shown in the 2nd row of Tab.~\ref{tab:efficiency}. Then, we explore two acceleration strategies for the DDPM denoiser: the widely used DDIM denoiser \cite{song2020denoising} (3rd and 4th rows) and the one-step TCD denoiser \cite{zheng2024trajectory} (5th row). Surprisingly, we find that the TCD denoiser largely preserves the segmentation performance of the DDPM denoiser while significantly improving efficiency, boosting FPS from 0.006 to 3.09. Furthermore, we fine-tune the VAE (6th row), which improves the model's ability to adapt to segmentation of high-resolution images without compromising efficiency. Finally, we compare the proposed \ourmodel~to the runner-up model MVANet \cite{yu2024multi}, our model demonstrates significant advantages in both segmentation performance and inference speed, as shown in the 1st and 6th rows of Tab.~\ref{tab:efficiency}.


\section{Conclusion}
\label{sec:con}

We have presented a model named \ourmodel~that reformulates the DIS task within a generative diffusion framework by extending a single stable diffusion model into macro and micro modes, thus achieving controllable DIS in two aspects. In the macro mode, an LS strategy is proposed that enables the generation of segmentation results under the control of language prompts. In the micro mode, an WR strategy is designed to enable unlimited detail optimization in controllable windows at variable scales on high-resolution images. The two modes are integrated into a single network through a switcher, enabling collaborative enhancement during training and seamless mode-switching during inference. Extensive experiments on the DIS5K dataset demonstrate that our \ourmodel~significantly outperforms SOTA methods across all the metrics.


\section*{Acknowledgement}
This work was supported by the National Natural Science Foundation of China (NO.62376189 \& NO.62476143). We express our sincere gratitude to Qi Ma (Nankai University) and Jingyi Liu (Keio University) for their constructive discussions.

{
    \small
    \bibliographystyle{ieeenat_fullname}
    \bibliography{main}
}

\newpage

\appendix
\clearpage
\setcounter{page}{1}

\onecolumn

\begin{center}
\large \textbf{LawDIS: Language-Window-based \\Controllable Dichotomous Image Segmentation}

\vspace{4pt}
\large Supplementary Material

\begin{figure*}[ht!]
\centering
\begin{overpic}[width=0.94\textwidth]{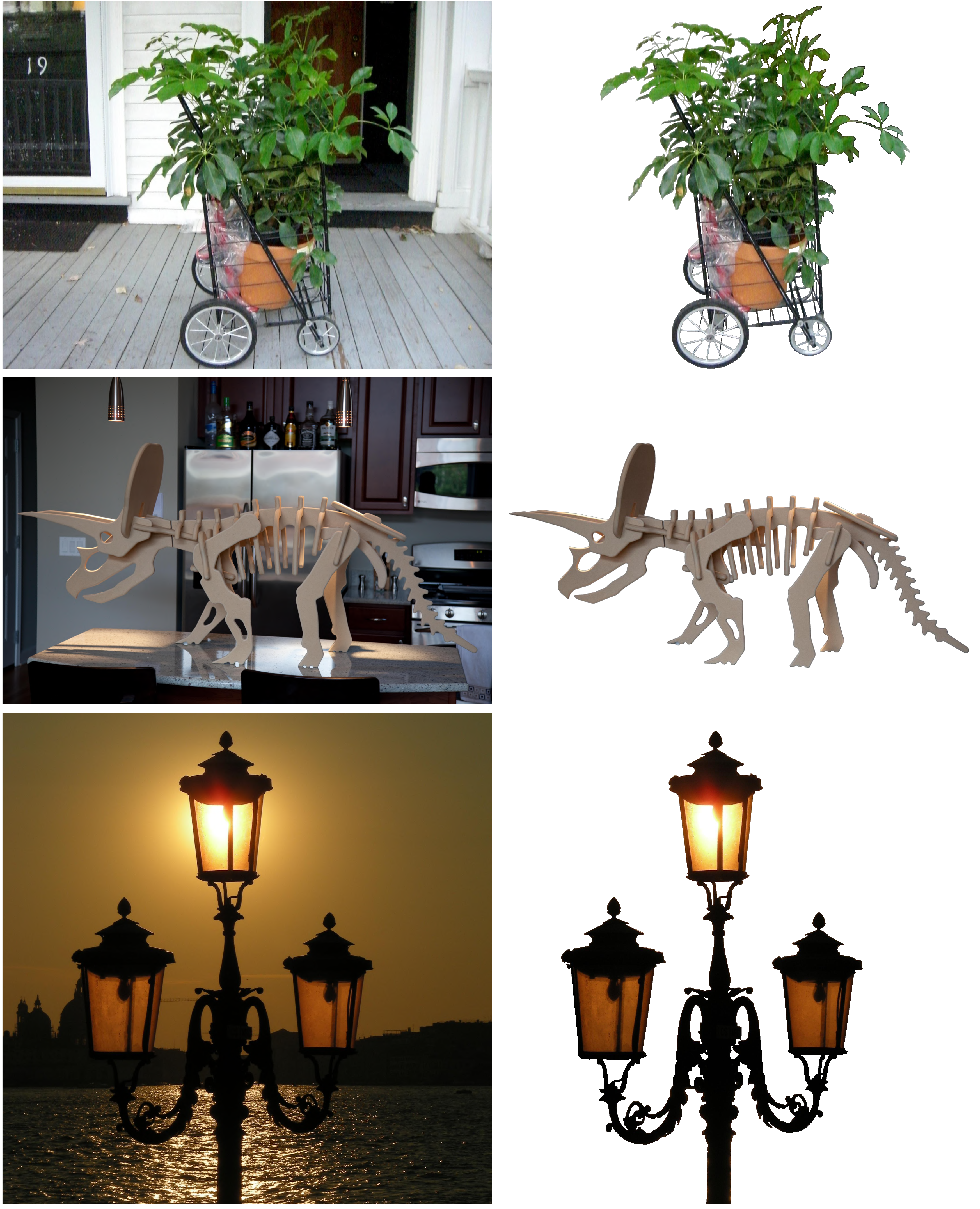}
\end{overpic}
\vspace*{-8pt}
\caption{Comparisons between the original images and their backgrounds-removed correspondences generated by our \ourmodel.
}
\vspace*{-8pt}
\label{fig:appendix_2}
\end{figure*}
\end{center}

\twocolumn

\section{Implementation Details}
\hypertarget{supplementary_anchor}{}
\subsection{Annotation Pipeline of Language Prompt}

We designed a semi-automated pipeline with three steps to generate a language prompt that accurately describes the foreground region of each image. Our pipeline effectively produces language descriptions that semantically align with each image's ground-truth (GT) mask, offering a high-quality foundation for building \ourmodel.

\begin{figure*}[t!]

\centering
\begin{overpic}[width=\textwidth]{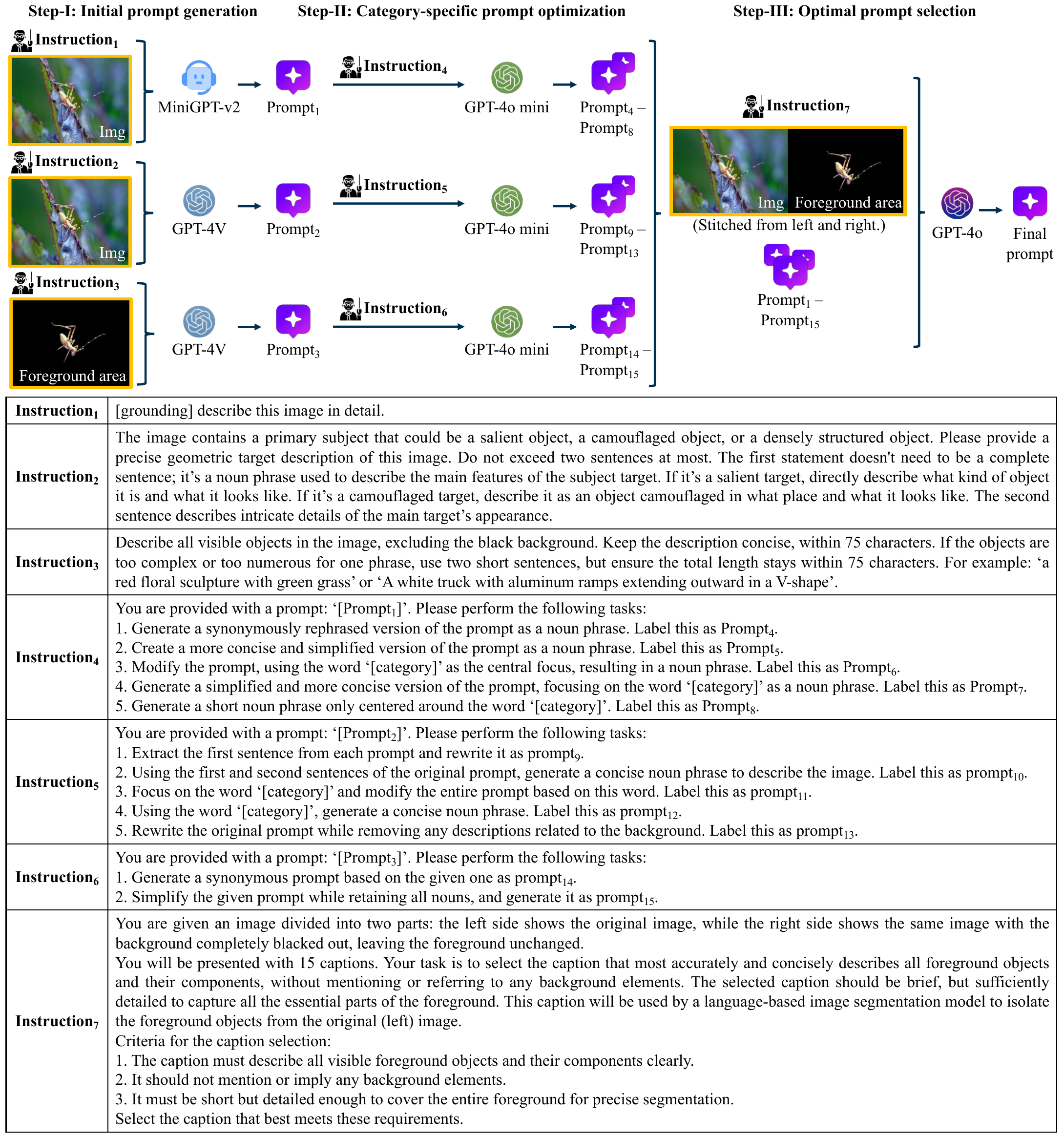}

\end{overpic}
\vspace*{-11pt}
\caption{Annotation pipeline of language prompt.
}

\vspace*{-10pt}
\label{fig:language_prompt}
\end{figure*}

\noindent\textbf{Step-I: Initial prompt generation.} This step generates three sets of descriptive captions for each image from the DIS5K dataset \cite{qin2022highly}. Specifically, as illustrated in Step-I of Fig.~\ref{fig:language_prompt}, we meticulously crafted Instruction$_{1}$ - Instruction$_{3}$ to command the generation of each prompt. We employ two multimodal chatbots, namely MiniGPT-v2 \cite{chen2023minigpt} and GPT-4V \cite{achiam2023gpt}, to generate language descriptions based on the entire image, designated Prompt$_{1}$ and Prompt$_{2}$, respectively. Furthermore, under the guidance of Instruction$_{3}$, GPT-4V generates Prompt$_{3}$, focusing solely on the foreground area\footnote{The foreground area is obtained by element-wise multiplication of the image and the GT mask.}.

\noindent\textbf{Step-II: Category-specific prompt optimization.} This step optimizes the three above generated prompts by integrating category-specific instructions. Specifically, in Step-II of Fig.~\ref{fig:language_prompt}, we design Instruction$_{4}$ - Instruction$_{6}$ to optimize the prompts generated in Step-I by incorporating the foreground object categories provided by the DIS5K dataset for each image. This optimization includes creating versions with synonymous expressions, more concise formulations, and category-specific descriptions. Using this instruction framework, we guide the GPT-4o mini \cite{achiam2023gpt} chatbot to generate more language descriptions, generating 12 new prompts: Prompt$_{4}$ to Prompt$_{8}$ derived from Prompt$_{1}$, Prompt$_{9}$ to Prompt$_{13}$ derived from Prompt$_{2}$, and Prompt$_{14}$ to Prompt$_{15}$ derived from Prompt$_{3}$. This not only broadens the descriptive aspects but also enhances the coherence between the prompts and the category-specific content.

\noindent\textbf{Step-III: Optimal prompt selection.} As illustrated in Step-III of Fig.~\ref{fig:language_prompt}, we stitch the original image and the foreground area side by side to form a single image, and use GPT-4o~\cite{achiam2023gpt} to select the optimal prompt from the 15 prompts generated earlier (\ie, Prompt$_{1}$ to Prompt$_{15}$), following the carefully designed Instruction$_{7}$.

\hypertarget{supplementary_window_anchor}{}
\subsection{Window Selection Strategy}

In the process of window-controlled refinement, there are two pipelines to select the window: one is user-selected through clicking and dragging in the semi-automated refinement pipeline, and the other is automatically generated in the fully automated refinement pipeline. Therefore, we design a window selection strategy that can simulate the user's window selection process in semi-automatic refinement pipelines to facilitate our experimental validation, while also enabling automatic window generation in fully automated refinment pipelines. Algorithm~\ref{algorithm: bounding-box} illustrates the pytorch-like pseudocode for the proposed strategy.

Specifically, in semi-automated pipeline, to simulate the process of users manually clicking to set windows, we take GT $s$ as input and generate windows to be refined around the edges of the foreground objects in GT. This aligns with real-world applications where the edges of foreground objects often require the most refinement. In fully automated pipeline, we use the initial language-controlled segmentation map $\hat{s }$ as the input for the window selection strategy, allowing the selection of windows around the object boundaries in the initial segmentation results. Notably, in practical applications, the windows in the semi-automated pipeline are generated by user clicks and do not require execution of this window selection strategy, providing greater flexibility.

\section{More Qualitative Comparisons}

\begin{figure*}[t!]
\centering
\begin{overpic}[width=\textwidth]{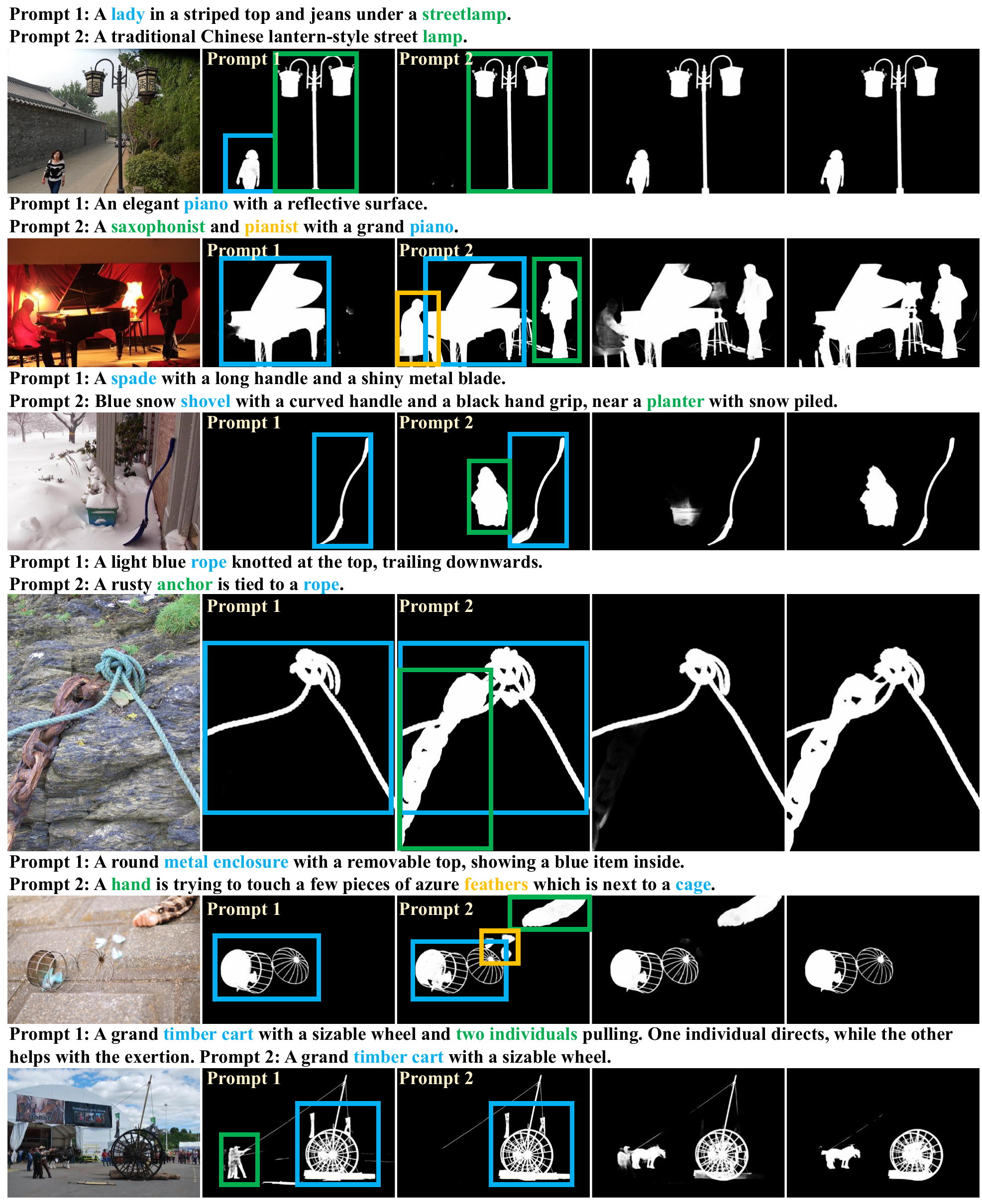}
    \put(5.5,-1){\small (a) Image}
    \put(29,-1){\small \textbf{(b) Ours-S}}
    \put(51.5,-1){\small (d) MVANet~\cite{yu2024multi}}
    \put(67.5,-1){\small (e) BiRefNet~\cite{zheng2024birefnet}}
\end{overpic}
\vspace*{-2pt}
\caption{Qualitative predictions under different prompt guidance. Our method flexibly segments various foreground objects based on prompts, while other methods that cannot accept prompts yield only a fixed result.
}
\vspace*{-1pt}
\label{fig:appendix_prompt}
\end{figure*}

\begin{figure*}[t!]
\centering
\begin{overpic}[width=\textwidth]{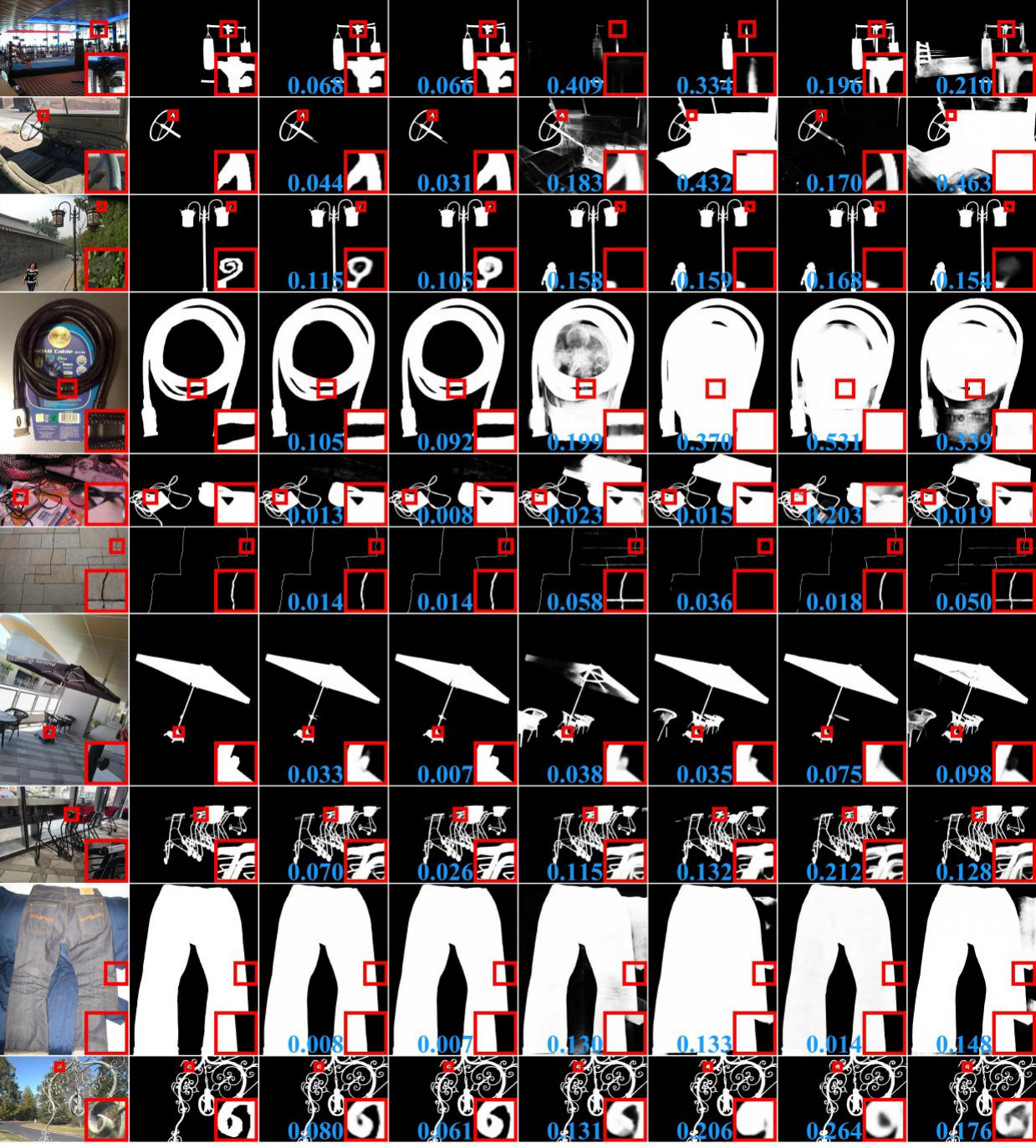}

    \put(3.5,-1){\small Image}
    \put(16,-1){\small GT}
    \put(26,-1){\small \textbf{Ours-S}}
    \put(37,-1){\small \textbf{Ours-R}}
    \put(46.3,-1){\small MVANet~\cite{yu2024multi}}
    \put(57,-1){\small BiRefNet~\cite{zheng2024birefnet}}
    \put(67.5,-1){\small GenPercept~\cite{xu2024diffusion}}
    \put(78.8,-1){\small InSPyReNet~\cite{kim2022revisiting}}

\end{overpic}
\vspace*{-13pt}
\caption{
Qualitative comparison of our model with four leading models. Local zoomed-in patches are evaluated with $\mathcal{M}$ score for clarity.
}
\vspace*{-8pt}
\label{fig:appendix_qua_com_VD}
\end{figure*}
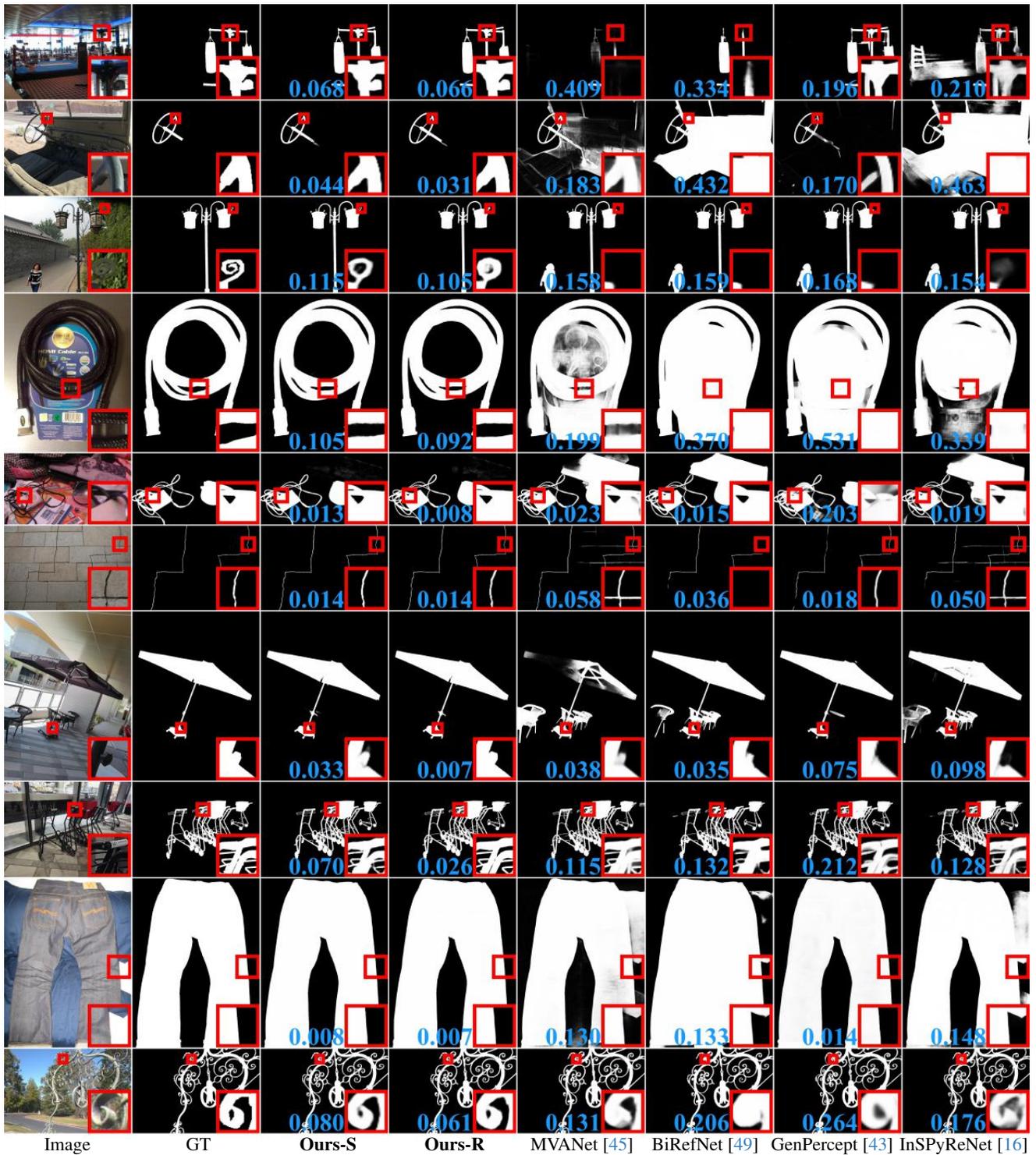

This section presents additional visual results. Fig.~\ref{fig:appendix_prompt} illustrates the qualitative predictions with different prompts. Our approach adaptively segments the corresponding foreground objects based on user-defined language prompts, whereas the other two SOTA methods~\cite{yu2024multi,zheng2024birefnet}, which lack prompt support, yield a fixed result. Moreover, Fig.~\ref{fig:appendix_qua_com_VD} presents qualitative comparisons between our method and the four top existing DIS models, further demonstrating its superior performance.

\section{Training cost \& model complexity}
As seen in Tab. \ref{tab:trainingcost}, LawDIS-S surpasses transformer-based MVANet \cite{yu2024multi} by 2.1\%, requiring fewer (36K) training iterations and achieving a faster runtime of 0.326s/image. Moreover, we select diffusion-based DiffDIS \cite{DiffDIS} for a fair comparison. Our LawDIS-S (macro mode w/o post-refinement) still has the upper hand in performance (0.925~\textit{vs.}~0.908) and inference speed (0.326s~\textit{vs.}~0.846s), despite comparable model parameters. All three models can run on an NVIDIA 4090 card, with user-friendly video memory consumption.

\begin{table}[h]
\centering
\resizebox*{\linewidth}{!}{
\begin{tabular}{l|lccl|cc}
\hline
Model            & Type                & Parameters &Memory    & Iter.             & Time $\downarrow$ & $F_\beta^{max}\uparrow$         \\ \hline
MVANet \cite{yu2024multi}             & Transformer                 & 369M   & 10.00G       &240K           & 0.714s    & 0.904   \\
DiffDIS \cite{DiffDIS}           & Diffusion                 & 3480M   &18.38G     & 67.5K           & 0.846s   & 0.908   \\
\rowcolor[HTML]{f5f5f5}
LawDIS-S            & Diffusion & 3537M  &17.12G     &\textbf{36K}            & \textbf{0.326s}    & \textbf{0.925}  \\ 
\hline
\end{tabular}
}
\caption{Comparison of training cost and model complexity.}
\label{tab:trainingcost}
\end{table}

\section{User study of controllability}

To evaluate user satisfaction with controllability, ten users participated in a rating study involving ten visual cases predicted in macro and micro modes, where a score of 5 means the best controllability. The results are shown in Tab. \ref{tab:controllability}.

\begin{table*}[ht]
\footnotesize
\renewcommand{\arraystretch}{1}
\renewcommand{\tabcolsep}{1.0mm}
\begin{tabular}{c|cccccccccc|c}
\hline
 &
  CASE\#1 &
  CASE\#2 &
  CASE\#3 &
  CASE\#4 &
  CASE\#5 &
  CASE\#6 &
  CASE\#7 &
  CASE\#8 &
  CASE\#9 &
  CASE\#10 &
  \textbf{Avg.   (macro/micro)} \\ \hline
USER\#01 & 5 / 5 & 5 / 5 & 5 / 5 & 5 / 5 & 5 / 5 & 5 / 4 & 5 / 5 & 5 / 5 & 5 / 5 & 5 / 5 & 5 / 4.9   \\
USER\#02 & 5 / 5 & 5 / 5 & 5 / 5 & 5 / 5 & 5 / 5 & 5 / 5 & 5 / 5 & 5 / 5 & 5 / 5 & 5 / 5 & 5 / 5     \\
USER\#03 & 5 / 4 & 4 / 5 & 5 / 5 & 5 / 5 & 5 / 5 & 5 / 5 & 5 / 5 & 5 / 4 & 5 / 5 & 5 / 4 & 4.9 / 4.7 \\
USER\#04 & 5 / 5 & 5 / 5 & 5 / 5 & 5 / 5 & 5 / 5 & 5 / 5 & 5 / 5 & 5 / 5 & 5 / 5 & 5 / 5 & 5 / 5     \\
USER\#05 & 4 / 5 & 5 / 4 & 5 / 4 & 5 / 5 & 5 / 5 & 5 / 5 & 4 / 5 & 5 / 5 & 5 / 5 & 5 / 5 & 4.8 / 4.8 \\
USER\#06 & 5 / 5 & 5 / 5 & 5 / 5 & 5 / 5 & 5 / 5 & 5 / 5 & 5 / 5 & 5 / 5 & 5 / 5 & 5 / 5 & 5 / 5     \\
USER\#07 & 5 / 5 & 5 / 5 & 5 / 5 & 5 / 5 & 5 / 5 & 5 / 5 & 5 / 5 & 5 / 5 & 5 / 5 & 5 / 5 & 5 / 5     \\
USER\#08 & 5 / 5 & 5 / 5 & 5 / 5 & 5 / 5 & 5 / 5 & 5 / 5 & 5 / 5 & 5 / 5 & 5 / 5 & 5 / 5 & 5 / 5     \\
USER\#09 & 5 / 5 & 5 / 5 & 5 / 5 & 5 / 5 & 5 / 5 & 5 / 5 & 5 / 5 & 5 / 5 & 5 / 5 & 5 / 5 & 5 / 5     \\
USER\#10 & 4 / 5 & 5 / 5 & 4 / 5 & 5 / 5 & 5 / 5 & 5 / 5 & 5 / 5 & 5 / 5 & 5 / 5 & 5 / 4 & 4.8 / 4.9 \\ \hline
\textbf{Avg.   (macro/micro)} &
  4.8   / 4.9 &
  4.9   / 4.9 &
  4.9   / 4.9 &
  5 / 5 &
  5 / 5 &
  5 / 4.9 &
  4.9 / 5 &
  5 / 4.9 &
  5 / 5 &
  5   / 4.8 &
  \textbf{4.95 / 4.93} \\ \hline
\end{tabular}
\vspace*{-8pt}
\caption{User study of controllability in macro and micro modes.}
\label{tab:controllability}
\end{table*}

\section{Discussion}
LawDIS is the first framework for the \textit{controllable} DIS task. Our innovation lies in repurposing the switcher to dynamically control generation at different levels of granularity, enabling a seamless transition between macro and micro parts. To achieve this, we designed new loss functions and a tailored training process. Compared to DPM \cite{lee2024exploiting}, which is an LDM with a DDPM scheduler designed for general segmentation, our method introduces specific mechanisms—such as the mode switcher and VAE fine-tuning—to better handle fine structures. Moreover, compared to MbG \cite{wang2024MG}, a two-stage diffusion model designed for human matting, our mode switcher connects macro segmentation with micro refinement during training, adaptively unlocking their synergy to enhance performance. Importantly, MbG-stage2 requires multiple patch-level refinements, each involving 50 time-consuming diffusion steps, which further highlights the efficiency advantage of our inference.

\section{Applications}
Due to its capability of achieving high-precision segmentation of foreground objects at high resolutions, our LawDIS enables extensive application across a variety of scenarios. Fig.~\ref{fig:appendix_2} shows the comparison between several original images and their corresponding background-removed versions. As can be seen, compared with the original image, the background-removed image shows higher aesthetic values and good usability, which can even be directly used for 3D modeling, augmented reality (AR), and still image animation. The corresponding demonstration videos (\texttt{app1-3D.mp4}, \texttt{app2-AR.mp4}, and \texttt{app3-Still-Image-Animation.mp4}) are available on this \href{https://space.bilibili.com/406752398/lists}{website}.


\begin{algorithm*}[!htb]

	\renewcommand{\algorithmicrequire}{\textbf{Input:}}
	\renewcommand{\algorithmicensure}{\textbf{Output:}}
	\caption{Window selection strategy (PyTorch-like pseudocode)}
	\label{algorithm: bounding-box}
	\begin{algorithmic}[0]
        \STATE \CommentGreen{~ $src$: source. In semi-automated pipeline: $src = GT s$; in fully automated pipeline: $src$ = initial segmentation map $\hat{s}$.} \\
        \CommentGreen{~ $(h_p, w_p)$: window size, defaulting to $(1024, 1024)$.} \\
        \CommentGreen{~ $K$: maximum number of selection iterations, defaulting to  $50$.}  \\
        \CommentGreen{~ $(\tau_1, \tau_2)$: the thresholds for edge strength in Canny edge detection, defaulting to $(50,150)$.}  \\
        \CommentGreen{~ $\tau_{\text{co}}$: the pixel connectivity to be considered when finding connected components, defaulting to $8$.}  \\
        \CommentGreen{~ $(\tau_{\text{white-pixel}},\tau_{\text{black-pixel}})$: the threshold values used to classify pixels as white or black, defaulting to $(250,5)$.}  \\
        \CommentGreen{~ $(\tau_{\text{white-region}},\tau_{\text{black-region}})$: the threshold values for the proportion of white and black regions, defaulting to $(0.9,0.95)$.} 
        \newline

        \texttt{i = 0, $ \mathcal{P}$ = [] } \CommentGreen{Initialize an iteration count and an empty list of window coordinates.}\\
        
        $ ~$\\
   
        \texttt{E = cv2.Canny(src, $\tau_1$, $\tau_2$)} \CommentGreen{Perform edge detection.} \\

        \CommentGreen{Compute connected components. $stats$ contains bounding box info and area, $centroids$ contains the center points.}\\
        \texttt{stats, centroids = cv2.connectedComponentsWithStats(E, $\tau_{\text{co}})$ } \\

        \texttt{stats = stats[1:], centroids = centroids[1:] } \CommentGreen{Remove background components.}\\

        \CommentGreen{Compute the maximum side length for each connected component.}\\
        \texttt{side\_lengths = max(stats[:, cv2.CC\_STAT\_WIDTH], stats[:, cv2.CC\_STAT\_HEIGHT]) }  \\
        
        $ ~$\\
        \CommentGreen{Iterate until there are no more components or maximum iterations are reached.}\\
        \texttt{\textbf{while} len(side\_lengths) > 0 \textbf{ and } i < K:} \\

            \hspace{10mm}\texttt{j = argmax(side\_lengths),  ($x_c$, $y_c$) = centroids[j]}   \CommentGreen{Find the largest connected component.}\\

            \hspace{10mm}\CommentGreen{Create a window centered at the center points of the selected connected component.}\\
             \hspace{10mm}\texttt{$x_1$ = max($x_c$ - $w_p$/2, 0),   $y_1$ = max($y_c$ - $h_p$/2, 0)} \\
             
             \hspace{10mm}\texttt{$x_2$ = min($x_c$ + $w_p$/2, W),   $y_2$ = min($y_c$ + $h_p$/2, H)} \\

             \hspace{10mm}\texttt{R = src[$y_1$:$y_2$, $x_1$:$x_2$]} \CommentGreen{Extract the window region.}\\

            \hspace{10mm}\CommentGreen{Compute the black and white pixel ratios.}\\
             \hspace{10mm}\texttt{white\_ratio = Count(R > $\tau_{\text{white-pixel}}$) / Size(R)} \\
             \hspace{10mm}\texttt{black\_ratio = Count(R > $\tau_{\text{black-pixel}}$) / Size(R)} \\
             $ ~$\\
            \hspace{10mm}\CommentGreen{If the proportion of white or black pixels exceeds the set threshold, ignore this window.} \\
             \hspace{10mm}\texttt{\textbf{if} white\_ratio > $\tau_{\text{white-region}}$ 
             \textbf{or} black\_ratio > $\tau_{\text{black-region}}$:}

            \hspace{20mm}\CommentGreen{Update the edge map by masking the current region.} \\
             \hspace{20mm}\texttt{M = \text{zeros\_like}(E)} \CommentGreen{Create a mask of the same size as the edge map.}\\
             \hspace{20mm}\texttt{M[$y_1$:$y_2$, $x_1$:$x_2$] = E[$y_1$:$y_2$, $x_1$:$x_2$]} \CommentGreen{Set the mask region to the current window.}\\
             \hspace{20mm}\texttt{E = BitwiseAnd(E, BitwiseNot(M))} \CommentGreen{Update the edge map by removing the masked region.}\\

            $ ~$\\
             \hspace{20mm}\CommentGreen{Recompute connected components.} \\
             \hspace{20mm}\texttt{stats, centroids = cv2.connectedComponentsWithStats(E,$\tau_{\text{co}}$)[1:]}    \\
             
             \hspace{20mm}\texttt{\textbf{continue}}

            $ ~$\\
             \hspace{10mm} \texttt{$\mathcal{P}$ = $\mathcal{P}$.append((($x_1$, $y_1$), ($x_2$, $y_2$)))} \CommentGreen{Record the window coordinates.} \\

             \hspace{10mm}\CommentGreen{Update the edge map by masking the current region.} \\
             \hspace{10mm}\texttt{M = \text{zeros\_like}(E)} \\
             \hspace{10mm}\texttt{M[$y_1$:$y_2$, $x_1$:$x_2$] = E[$y_1$:$y_2$, $x_1$:$x_2$]} \\
             \hspace{10mm}\texttt{E = BitwiseAnd(E, BitwiseNot(M))} \\

            $ ~$\\
             \hspace{10mm}\CommentGreen{Recompute connected components.} \\
             \hspace{10mm}\texttt{stats, centroids = cv2.connectedComponentsWithStats(E, $\tau_{\text{co}}$)[1:]}  \\
             $ ~$\\
             \hspace{10mm}\texttt{i = i + 1} \CommentGreen{Increment the iteration count.} \\
        $ ~$\\
         \texttt{\textbf{return} $ \mathcal{P}$} \CommentGreen{Return the list of window coordinates.} 
             
	\end{algorithmic}
\end{algorithm*}

\end{document}